%% file: arxiv.tex
\documentclass[10pt,twocolumn,letterpaper]{article}

\usepackage{iccv}
\usepackage{times}
\usepackage{epsfig}
\usepackage{graphicx}
\usepackage{amsmath}
\usepackage{amssymb}
\usepackage{subcaption}
\usepackage{blindtext}
\usepackage{booktabs}  
\usepackage{multirow}
\usepackage{enumitem}
\usepackage{pifont}
\usepackage{tabulary,multirow,overpic,xcolor,subfloat}
\usepackage{colortbl}
\usepackage[pagebackref=true, breaklinks=true, letterpaper=true, colorlinks, citecolor=citecolor, bookmarks=false]{hyperref}
\usepackage{cleveref}
\usepackage{algorithm}
\usepackage{algorithmic}
\usepackage{listings}
\usepackage{diagbox}
\usepackage{mathtools}

\definecolor{citecolor}{HTML}{0071BC}

\input{definition.tex}
\iccvfinalcopy %

\def\iccvPaperID{****} %

\begin{document}

\title{Denoising Diffusion Semantic Segmentation with Mask Prior Modeling}

\author{
Zeqiang Lai\textsuperscript{1,4$\dagger *$} 
\quad Yuchen Duan\textsuperscript{3,4 *}
\quad Jifeng Dai\textsuperscript{2,4}
\quad Ziheng Li\textsuperscript{2}
\quad Ying Fu\textsuperscript{1}
\\
\quad Hongsheng Li\textsuperscript{3}
\quad Yu Qiao\textsuperscript{4}
\quad Wenhai Wang\textsuperscript{4}
\\
\textsuperscript{1}{Beijing Institute of Technology}
\quad \textsuperscript{2}{Tsinghua University}
\quad \textsuperscript{3}{CUHK}
\quad \textsuperscript{4}{Shanghai AI Laboraotry}
\\
\\
\small{Code: \url{https://github.com/OpenGVLab/DDPS}}
\\
\small{Demo: \url{https://github.com/OpenGVLab/InternGPT}}
}

\maketitle
\ificcvfinal\thispagestyle{empty}\fi

\blfootnote{$^\dagger$Work done during an internship at Shanghai AI Laboraotry.}
\blfootnote{* Equal Contribution.}

\input{sections/abstract.tex}
\input{sections/introduction.tex}
\input{sections/related_works.tex}

\input{sections/method.tex}

\input{sections/experiment.tex}

\input{sections/conclusion.tex}

{\small
\bibliographystyle{ieee_fullname}
\bibliography{egbib}
}

\newpage
\clearpage
\input{sections/appendix.tex}

\end{document}

%% file: definition.tex
\definecolor{Gray}{rgb}{0.5,0.5,0.5}
\definecolor{darkblue}{rgb}{0,0,0.7}
\definecolor{orange}{rgb}{1,.5,0} %
\definecolor{red}{rgb}{1,0,0} %

\newcommand{\figref}[1]{Fig.~\ref{#1}}
\newcommand{\tabref}[1]{Table~\ref{tab:#1}}

\newcommand\blfootnote[1]{%
  \begingroup
  \renewcommand\thefootnote{}\footnote{#1}%
  \addtocounter{footnote}{-1}%
  \endgroup
}

\newcommand{\bx}{\boldsymbol{x}}

\newcommand{\bs}[1]{\boldsymbol{#1}}

\newcommand{\one}{$1^{\text{st}}$ step}

\newcommand{\twenty}{$20^{\text{th}}$ step}

\definecolor{graycolor}{rgb}{0.95,0.95,0.95}

\DeclarePairedDelimiterX{\infdivx}[2]{(}{)}{%
  #1\;\delimsize|\delimsize|\;#2%
}
\newcommand{\kld}[2]{\ensuremath{D_{KL}\infdivx{#1}{#2}}\xspace}

%% file: sections/abstract.tex
\begin{abstract}
The evolution of semantic segmentation has long been dominated by learning more discriminative image representations for classifying each pixel. Despite the prominent advancements, the priors of segmentation masks themselves, \eg, geometric and semantic constraints, are still under-explored. In this paper, we propose to ameliorate the semantic segmentation quality of existing discriminative approaches with a mask prior modeled by a recently-developed denoising diffusion generative model. Beginning with a unified architecture that adapts diffusion models for mask prior modeling, we focus this work on a specific instantiation with discrete diffusion and identify a variety of key design choices for its successful application. Our exploratory analysis revealed several important findings, including: (1) a simple integration of diffusion models into semantic segmentation is not sufficient, and a poorly-designed diffusion process might lead to degradation in segmentation performance; (2) during the training, the object to which noise is added is more important than the type of noise; (3) during the inference, the strict diffusion denoising scheme may not be essential and can be relaxed to a simpler scheme that even works better. We evaluate the proposed prior modeling with several off-the-shelf segmentors, and our experimental results on ADE20K and Cityscapes demonstrate that our approach could achieve competitively quantitative performance and more appealing visual quality.

\end{abstract}

%% file: sections/introduction.tex
\section{Introduction}

\begin{figure}[t]
\centering
\includegraphics[width=1\linewidth]{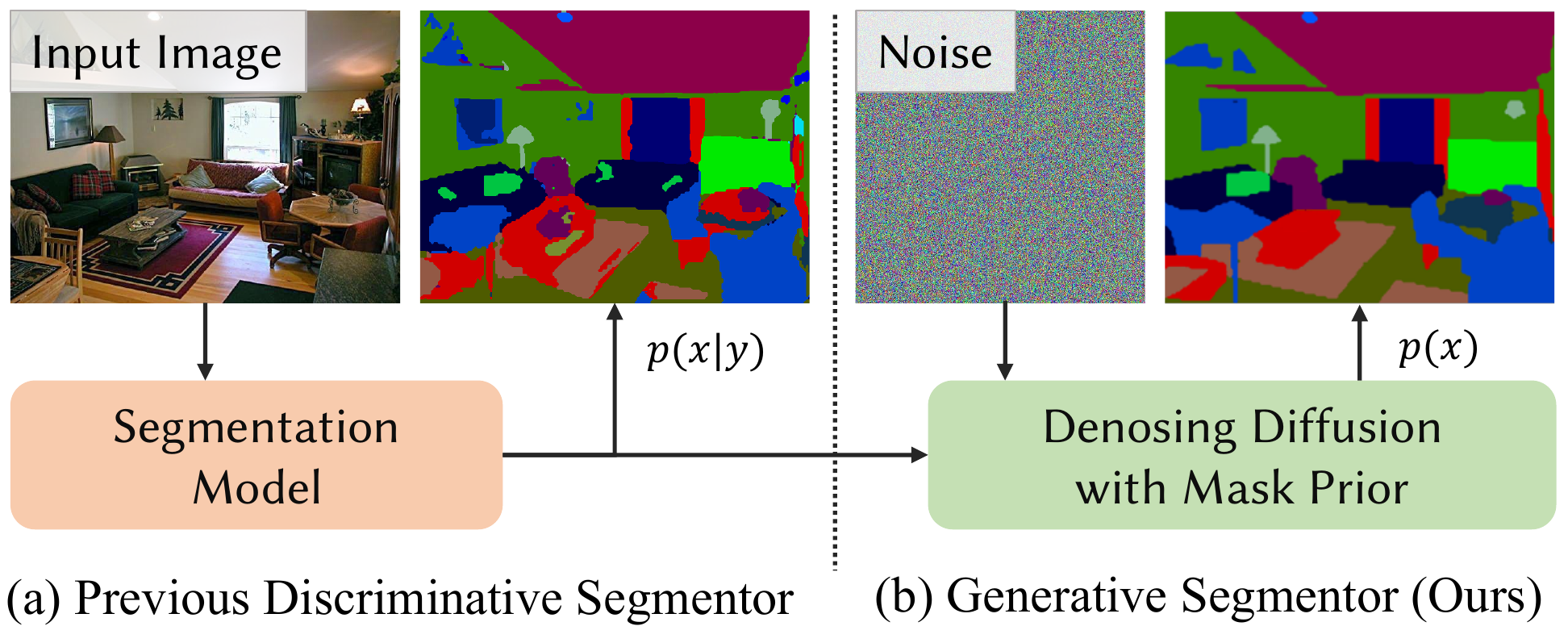}
\caption{
Conventional discriminative segmentation models (segmentors) typically learn elaborated image features for classifying pixels of an input image $y$, maximizing the posterior $p(x|y)$, but the intrinsic properties of the segmentation masks $x$ themselves are often overlooked. This work aims at bridging the gaps by drifting initial predictions towards the distribution matching the segmentation mask prior $p(x)$ with a denoising diffusion generative model, which is dubbed as \emph{denoising diffusion prior}. 
}
\label{fig:teaser}
\end{figure}

Semantic segmentation assigns category labels for every pixel of a given input image. As a representative visual recognition task, semantic segmentation supports a broader range of applications such as autonomous driving~\cite{feng2020deep,philion2020lift,li2022bevformer}, medical image analysis~\cite{ronneberger2015u,dong2021polyp,fan2020pranet}, and augmented reality~\cite{garcia2017review,azuma1997survey,billinghurst2015survey}. Although prominent advancements achieved since the prevalence of deep learning, challenges still exist for expanding the current paradigms~\cite{cheng2021per,long2015fully,he2017mask} to lift the label accuracy and visual coherence to a higher level.

Since the introduction of FCN~\cite{long2015fully}, the evolution of semantic segmentation has long been dominated by learning more discriminative image representations for classifying each pixel. 
Following this paradigm, image representation learning~\cite{dosovitskiy2020image,he2022masked,he2016deep} has been widely studied.
Benefiting from the modern backbone architectures~\cite{he2016deep,liu2022convnet, dosovitskiy2020image,liu2021swin}, engineering techniques~\cite{wightman2021resnet} and large-scale datasets~\cite{deng2009imagenet,schuhmann2022laion}, these models could generate remarkable informative image features for downstream tasks including semantic segmentation. To further elevate the performance, modern segmentation methods often adopt diverse post-processing pipelines, so-called segmentation heads~\cite{zhao2017pyramid,yu2020context,chen2014semantic, chen2017rethinking}, for enriching the vanilla backbone features with multi-scale or global contextual information.
Nevertheless, most of these methods, from the perspective of probability, can be unified as  
discriminative models,  which learn better image features for classifying the pixels of image $x$ into segmentation mask $y$, maximizing the posterior $p(x|y)$ as shown in Fig.~\ref{fig:teaser}(a). 

\begin{figure}[t]
\centering
\includegraphics[width=1\linewidth]{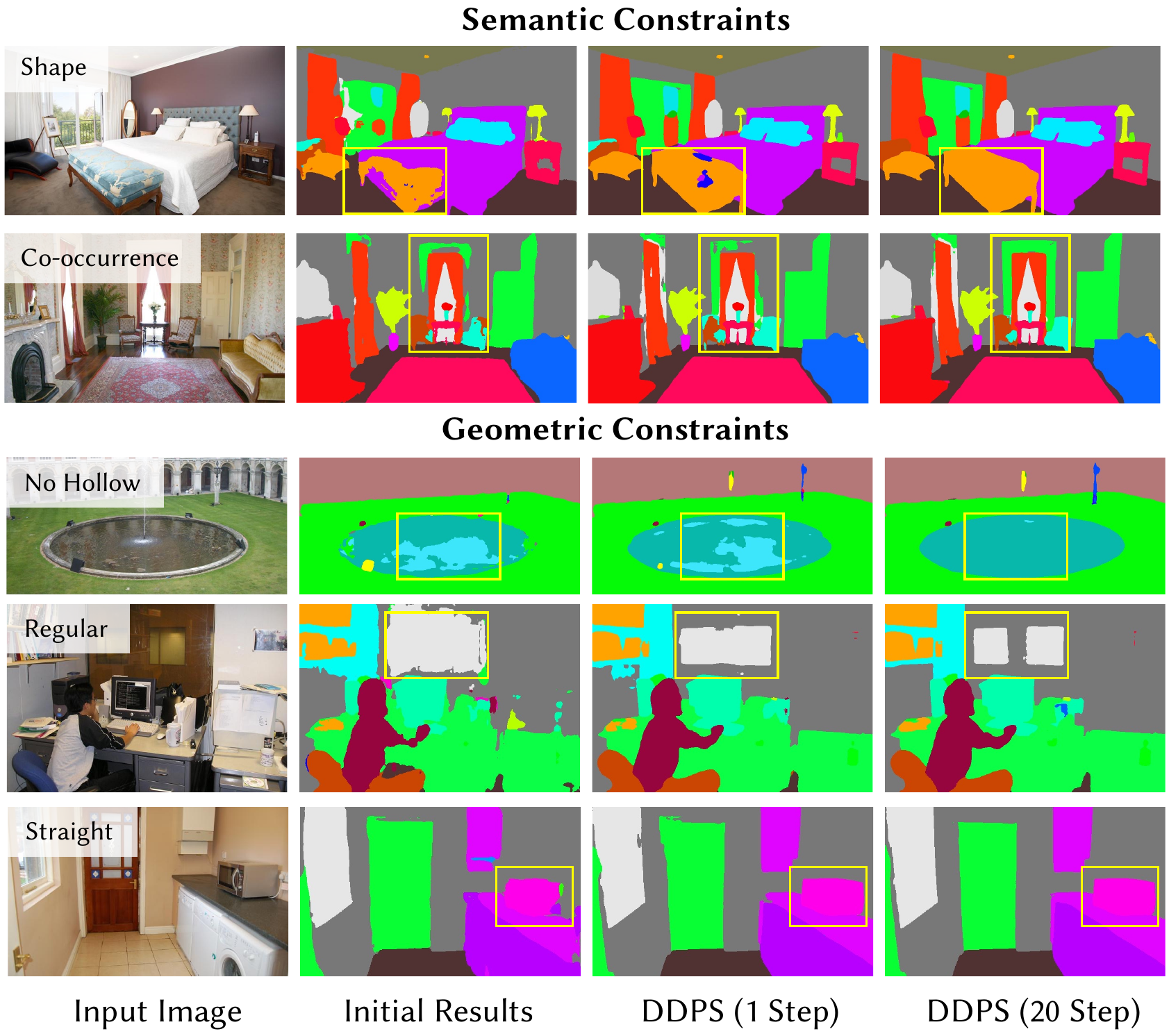}
\caption{
The prior knowledge of segmentation masks: semantic constraints (\eg, a specific shape of a class of objects, co-occurrence of certain classes of objects) and geometric constraints (\eg, no hollow, regular, straight).
}
\vspace{-5pt}
\label{fig:ob}
\end{figure}

Focusing on the feature learning of input image side, the previous prevalent paradigms~\cite{zhao2017pyramid,chen2017rethinking,xie2021segformer}, to some extent, inevitably suffer from the overlook or show limitations at capturing the prior information $p(x)$ of the output segmentation mask side, \eg, geometric and semantic constraints as shown in \figref{fig:ob}. 
Some previous works~\cite{zhao2017pyramid,yu2020context} have partly utilized the prior knowledge of segmentation masks. For example, PSPNet~\cite{zhao2017pyramid} identifies the constraints of label relationships, \eg, boats are more likely over water than cars. CPNet~\cite{yu2020context} explores the intra-class contextual dependencies, \ie, pixels of an object should have the same labels. 
However, these methods are still confined to feature learning with discriminative objectives. Direct modeling of segmentation mask prior for improving the results by maximizing the prior distribution $p(x)$, resembling the success paradigm of the image prior ~\cite{wang2022zero,kadkhodaie2020solving,wang2021towards} in low-level vision, is still under-explored.

The prior modeling for data of different modalities has been substantially advanced with the development of deep generative models~\cite{goodfellow2020generative,ho2020denoising,kingma2013auto,rezende2015variational,brown2020language}.
For semantic segmentation, Generative Adversarial Networks (GAN) ~\cite{goodfellow2020generative} are previously adopted for modeling the joint distribution of images and segmentation mask $p(x,y)$ ~\cite{li2021semantic} and imposing geometric constraints through adversarial training ~\cite{kim2021unsupervised}.  
However, these GAN-based approaches seldom yield more competitive performance than modern segmentors and their complex pipelines and training instabilities also hinder further developments.
Recently, denoising diffusion generative models ~\cite{ho2020denoising} have obtained unprecedented advancements in content generation tasks ~\cite{rombach2022high,villegas2022phenaki,poole2022dreamfusion}. Their simplicity and superior achievements spur a strong interest in adapting them for a broader range of applications ~\cite{luo2021diffusion,gong2022diffuseq}.  
Albeit the preliminary attempts on image segmentation ~\cite{baranchuk2021label,chen2022generalist,amit2021segdiff,wu2022medsegdiff} with continuous diffusion ~\cite{ho2020denoising}, from a generative perspective, it remains unclear how we can utilize the merits of diverse diffusion models ~\cite{ho2020denoising, austin2021structured} to model the prior of segmentation mask that benefits the semantic segmentation in general complex scenes.

In this work, we aim to ameliorate the semantic segmentation quality of existing discriminative approaches with a segmentation mask prior modeled by recently-developed denoising diffusion generative models ~\cite{ho2020denoising}, as shown in \figref{fig:teaser}(b). 
To achieve this, we first construct a unified architecture adapting Denoising Diffusion Prior modeling for semantic Segmentation (DDPS).
We conceptually divide the architecture into three parts to facilitate future exploration, including (1) a mask representation codec transforming the mask to make it more amenable to diffusion models, (2) a base segmentation model emitting the initial predictions, (3) a diffusion segmentation prior that iteratively refines the initial predictions towards the outcome that better matches the mask prior distribution.
Different from previous works ~\cite{chen2022generalist,wu2022medsegdiff} that focus on contiguous diffusion segmentation, \emph{this work
targets to explore a different possibility with discrete diffusion ~\cite{austin2021structured} and identify a variety of key design choices for its successful application.}

With our exploration, we identify several important findings as follows: (1) the integration of diffusion models and semantic segmentation must be carefully designed, as a poorly-designed diffusion process can result in degraded segmentation performance;
(2) the diffusion during the training $q(x_t|x_0)$ (constructing noisy samples $x_t$ for training) is significant, and this process is influenced more by the object to which noise is added, rather than the type of noise used;
(3) the previous strict diffusion during the inference $q(x_{t-1}|x_t,x_0)$ (deriving next state $x_{t-1}$ based on current state $x_{t}$ and clean estimation $x_{0}$) may not be essential and direct reuse of training diffusion scheme as $q(x_{t-1}|x_0)$ could simply work better. 
Following these findings, we evaluate the proposed prior modeling with several representative off-the-shelf segmentation methods \cite{xie2021segformer,deeplabv3plus2018}, and the experimental results on ADE20K~\cite{zhou2017scene} and Cityscapes~\cite{Cordts2016Cityscapes} demonstrate that our approach could achieve competitively quantitative performance and more appealing visual quality than previous discriminative segmentors~\cite{xie2021segformer,deeplabv3plus2018,zhao2017pyramid,yuan2020object}.
Notably, with our DDPS, the mIoU of Segformer-B2 on ADE20K can be improved from 46.80\% to 49.73\%, an improvement of over 3 points. 
We hope this investigation of the prior modeling of task outcomes (\eg, segmentation masks) could inspire further research that pushes the performance boundary even higher, for semantic segmentation and beyond.

%% file: sections/related_works.tex
\section{Related Works}

\textbf{Semantic Segmentation.} 
Ever since the introduction of the Full-Convolutional Network (FCN) \cite{long2015fully}, semantic segmentation has been conventionally regarded as a task of pixel-level classification.
Under this paradigm, most previous works share a common encoder-decoder architecture. The encoder typically comprises a backbone network \cite{he2016deep} and, optionally, a neck network \cite{zhao2017pyramid,zhao2018icnet}, \eg, feature pyramid network (FPN) \cite{zhao2017pyramid}. Learning more discriminative image representations have long been the central task of image encoder, which gives birth to various advanced backbone architectures, \eg, different convolutional neural networks \cite{he2016deep,liu2022convnet} and vision transformers \cite{dosovitskiy2020image,liu2021swin,xie2021segformer,zheng2021rethinking}, and novel training strategies, \eg, mask image modeling \cite{he2022masked,bao2021beit} and contrasted learning \cite{radford2021learning}. Unlike image classification, semantic segmentation usually requires more elaborated features for dense prediction. To this end, diverse segmentation decoders \cite{ronneberger2015u,chen2014semantic,zhao2017pyramid} are introduced with techniques, \eg, skip-connection \cite{ronneberger2015u}, dilate convolution \cite{chen2014semantic, chen2017rethinking}, and pyramid pooling \cite{zhao2017pyramid}, to enrich the encoder features with more high-resolution contextual information. This paradigm has dominated the research of semantic segmentation for a long time until the emergence of  
mask prediction paradigm \cite{zhang2021k,cheng2021per} that predicts a set of binary masks assigned to each class. Inspired by the object queries in DETR\cite{carion2020end}, representative works of this paradigm like MaskFormer \cite{cheng2021per} and K-Net \cite{zhang2021k} have achieved superior performance than those per-pixel classification ones while maintaining a unified architecture for different segmentation tasks. Recent works in Mask2Former \cite{cheng2022masked} and Mask-DINO again further advances the performance by refining the designs. Different from the previous paradigms, this work aims at exploring another approach that focuses more on the modeling of output segmentation masks instead of the input images.

\textbf{Denoising Diffusion Model.} 
Denoising Diffusion Probability Model (DDPM) \cite{ho2020denoising, song2019generative, sohl2015deep} is a recently developed generative model. 
Different from its precursors, e.g., Generative Adversarial Network (GAN) \cite{goodfellow2020generative}, Variational Autoencoder(VAE) \cite{kingma2013auto}, and normalizing flow \cite{rezende2015variational}, DDPM possesses a variety of advantages, e.g., training stability, simplicity, and convenience for downstream tasks \cite{saharia2022palette, wang2022zero}. 
Recent research has shown that diffusion models can generate artificial contents with unprecedented quality and diversity for different modalities, including images \cite{ramesh2022hierarchical, saharia2022photorealistic}, videos \cite{villegas2022phenaki,ho2022video}, audio \cite{kong2020diffwave}, and text \cite{li2022diffusion,gong2022diffuseq}. 
To model diverse modal data, a variety of diffusion models, \eg, DDPM \cite{ho2020denoising}, Discrete-Diffusion \cite{gu2022vector,austin2021structured}, Bit-Diffusion \cite{chen2022analog}, and Soft-Diffusion \cite{daras2022soft}, have been proposed by leveraging different continuous and discrete diffusion processes.  Most of the diffusion models apply for different problems with proper reformulation \cite{chen2022analog,dieleman2022continuous}, but the continuous diffusions \cite{ho2020denoising} are usually more prevalent for ordinal data, \eg, images and videos, while discrete diffusions \cite{austin2021structured, gu2022vector} for discrete data, \eg, text. For image segmentation, continuous diffusion has been explored in \cite{chen2022generalist, wu2022medsegdiff,wolleb2022diffusion}. In this work, we target to explore a different possibility with discrete diffusion \cite{gu2022vector}.

\textbf{Diffusion Models for Visual Perception.} Besides the substantial advancement in the realms of content generation, the diffusion models have also exhibited gradual promise within the domain of perception \cite{chen2022diffusiondet,gu2022diffusioninst,chen2022generalist, brempong2022denoising}. 
Early works in this field mostly focus on exploring latent representations of the diffusion model for zero-shot image segmentation \cite{baranchuk2021label, burgert2022peekaboo}, or apply diffusion models for medical image segmentation \cite{wu2022medsegdiff,wolleb2022diffusion}. 
Despite some progress achieved, these attempts only show results on limited applications. 
Recent Pix2Seq-D \cite{chen2022generalist} extends the bit-diffusion \cite{chen2022analog} for panoptic segmentation,  for the first time, in a broader context. Other than that, diffusion models for the query-based object detection \cite{carion2020end} and instance segmentation \cite{zhang2021k} are also explored in DiffusionDet \cite{chen2022diffusiondet} and DiffusionInst \cite{gu2022diffusioninst}. However, it still remains unclear if diffusion models could benefit semantic segmentation in a more general complex scenario.

%% file: sections/method.tex
\section{Denoising Diffusion Mask Prior Modeling}

\begin{figure*}[t]
\centering
\includegraphics[width=0.98\linewidth]{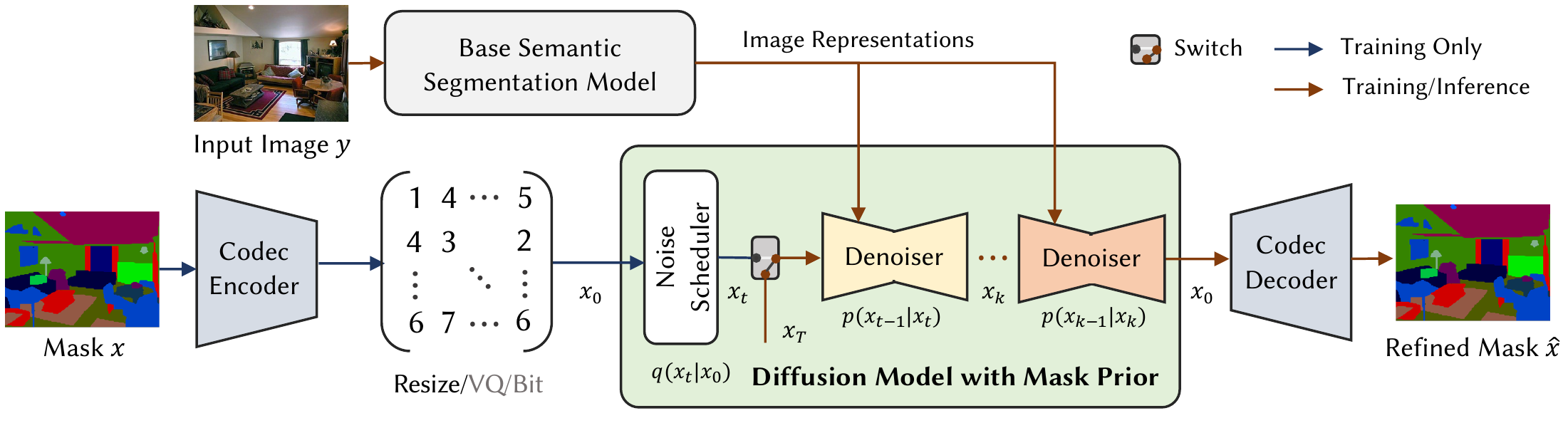}
\caption{\textbf{Architecture Overview.}
DDPS modularizes the diffusion segmentation with a two-stage pipeline and three major components. The first stage employs an off-the-shelf base segmentation model to produce the initial predictions that are refined by the second stage with a diffusion prior and a mask representation codec.
Please refer to the text for more details.
}
\label{fig:arch}
\end{figure*}

We attempt to ameliorate existing semantic segmentation paradigms ~\cite{long2015fully,cheng2021per} by introducing the prior modeling for segmentation masks. To achieve it, we adopt a two-stage formulation, where the first stage involves conventional discriminative segmentation models ~\cite{zhao2017pyramid,xie2021segformer} that directly model conditional probability $p(x|y)$, and the second stage refines the initial predictions by gradually drifting them towards outcomes with higher prior probability $p(x)$ so that the final results could better match the prior distribution of segmentation masks.

Instead of learning a black-box model that maps the initial predictions, 
we propose to model the second-stage refinement as an optimization problem where we progressively correct the predictions toward the final outcome with diffusion generative models~\cite{ho2020denoising}. 
In the following, we first illustrate some representative prior knowledge of segmentation masks, then specify a unified architecture, dubbed as DDPS, for adapting Denoising Diffusion Prior modeling for semantic Segmentation, and finally techniques used in the training and inference processes of our specific instantiation of DDPS with discrete diffusion ~\cite{austin2021structured}.

\subsection{Observations}

Similar to the image priors utilized for image restoration ~\cite{heide2016proximal}, such as the tendency for most image gradients to be close to zero (also known as total variation), the task of semantic segmentation of scenes and objects also entails some inherent prior knowledge and constraints. In the following, we detail two typical segmentation priors, namely geometric and semantic constraints.

\textbf{Semantic Constraints.}  Objects that occur in the same scenes typically correlated with each other. For example, the pair of airplane and sky is more likely to appear concurrently. Such relationships were usually implicitly modeled by incorporating more global contextual information in previous CNN-based methods ~\cite{zhao2017pyramid,yuan2020object}, but recent transformer-based methods ~\cite{zhang2021k,cheng2021per} also introduce a way with self-attention for performing interaction between categories, inspired by the learnable queries from object detection ~\cite{carion2020end}. 

\textbf{Geometric Constraints.} 
Natural objects and stuff typically exhibit inherent continuity, with few instances of discontinuous voids within their interior. Due to the semantic ambiguity, this has been one of the common errors of existing segmentation models and discussed in ~\cite{zhao2017pyramid,yu2020context}. Moreover, as shown in \figref{fig:ob}, artificial objects, \eg, buildings and furniture, often show notable geometric patterns like straight lines and smooth contours and boundaries. In a sense, imposing this prior knowledge can be somehow difficult to be guaranteed for existing methods without supervision on mask prior modeling.

\subsection{Overall Architecture}

\figref{fig:arch} shows the DDPS architecture. It mainly consists of three parts: i) a base segmentation model that gives the initial predictions in terms of image representations, ii) a diffusion segmentation prior that performs iterative refinement by drifting the initial results towards the outcome that better matches the mask prior distribution, and iii) a mask representation codec that transforms the segmentation mask so that it is amenable to the generative network processing.

\textbf{Base Segmentation Model.} 
DDPS follows a two-stage segmentation pipeline, where the first-stage base segmentation model can be any existing off-the-shelf segmentors ~\cite{zhao2017pyramid,xie2021segformer} that offer the initial segmentation results which are latterly refined by the diffusion segmentation prior model. The base segmentation model is fixed during the training and only runs once during the inference.  To retain more information, we adopt the high-level image representations before the last classifier layer instead of raw mask logits.

\textbf{Denoising Diffusion Mask Prior.} 
The key of DDPS is its second stage model with a learned denoising diffusion segmentation prior. By formulating the refinement as an optimization problem progressively solved by the diffusion prior, DDPS favors the alignment of distributions between initial mask predictions and real masks with explicit generative modeling.  
Though a variety of diffusion models ~\cite{ho2020denoising, chen2022analog, austin2021structured} are applicable, we focus this work on a specific instantiation with discrete diffusion ~\cite{austin2021structured}. In a nutshell, starting from a stationary state $x^T$, \eg, random mask or blank mask with all pixels belonging to a special $\text{[MASK]}$ category, the diffusion model with mask prior then iteratively denoises the current state $x^t$ to the next state $x^{t-1}$, conditioned on initial predictions of image representations. Following ~\cite{ho2020denoising}, we adopt a variant of U-Net ~\cite{ronneberger2015u} as denoiser. It is trained to predict the unknown clean mask $x_0$ for deriving $x_{t-1}$ instead of directly predicting $x_{t-1}$, which is also known as $x_0$ parameterization. 

\textbf{Mask Representation Codec.} We encode the segmentation mask into a representation that is more amenable to generative network processing. For example, we could quantize the segmentation mask with a VQ-VAE ~\cite{van2017neural} and performs discrete diffusion at the space defined by the VQ-codebook. 
Besides, the formulation in ~\cite{chen2022analog} can also be used to convert masks into analog bits with binary encoding.
Not surprisingly, DDPS unifies models with different codecs, \eg, Pix2Seq-D ~\cite{chen2022generalist} can be viewed as a special instantiation with a bit codec and Gaussian diffusion prior. 
In this work, we opt for the simplicity and the computational benefits of a simple resize codec and utilize discrete diffusion ~\cite{austin2021structured} that works directly on the segmentation mask.

\begin{figure}[t]
\centering
\includegraphics[width=1\linewidth]{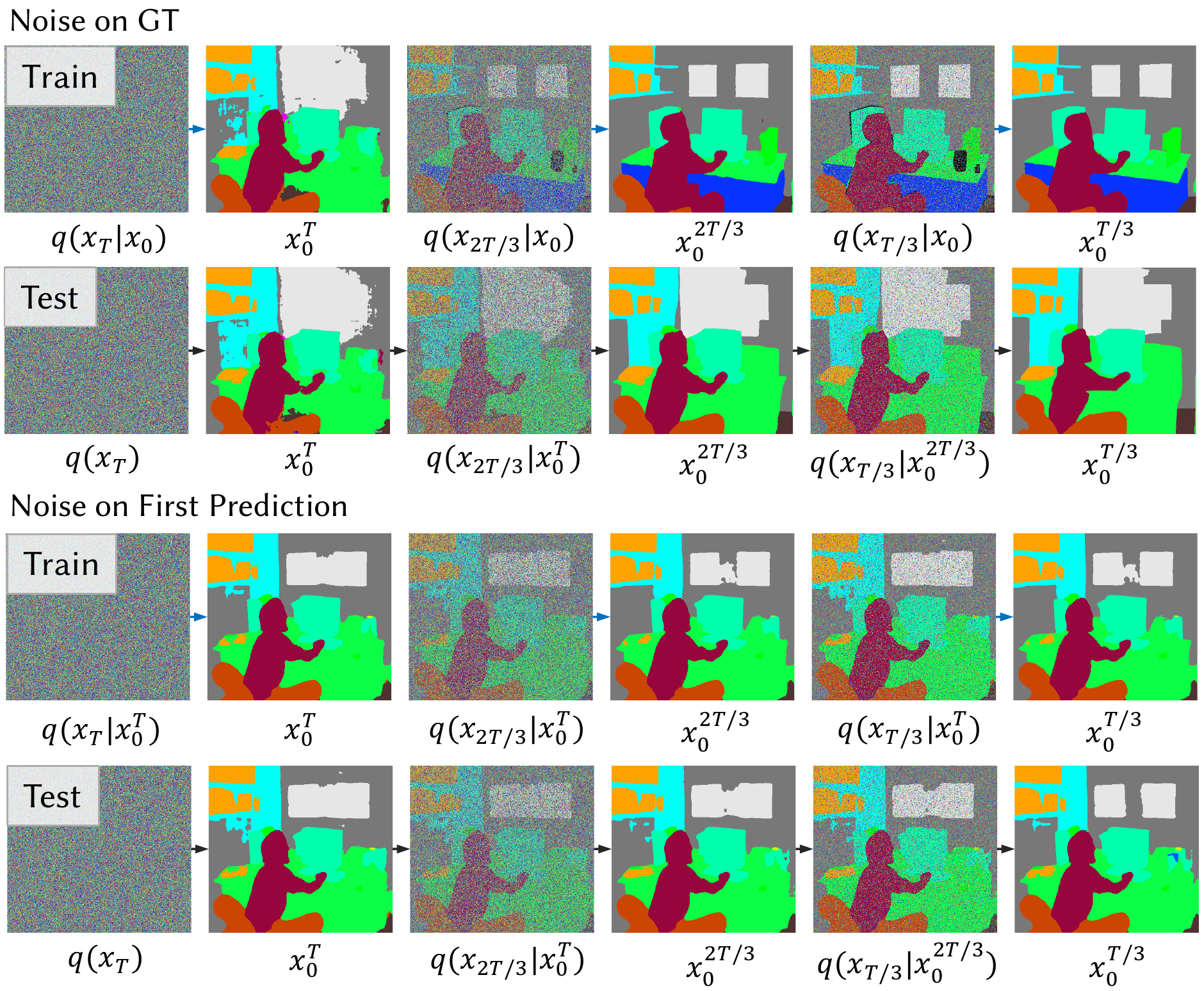}
\caption{The diffusion models are typically trained to reconstruct the ground truth (GT) given corrupted GT. However, the input of the diffusion model is its previous output instead of GT during the inference, which would
lead to a distribution gap that deteriorates the performance.
}
\label{fig:noise-target}
\end{figure}

\subsection{Training}
\label{sec:method:diffusion}

During the training, the diffusion model is trained to reconstruct the data $x_0$ given its corrupted version $x_t$ sampled from a diffusion process $q(x_t|x_0)$ with a random timestep $t$. Typically, both the diffusion target and reconstruct target are ground truth data ~\cite{ho2020denoising}. 
However, we find that such a choice would result in a distribution shift between training and inference for the input of the denoising diffusion segmentation model. 
Moreover, our further exploration suggests that the proper choice of diffusion target and training criterion is significant for the success of diffusion segmentation while the choice of diffusion transition is less important.

\textbf{Diffusion on First Prediction.}
The conventional diffusion processes ~\cite{ho2020denoising,austin2021structured} generate noisy samples by corrupting the ground-truth (GT) data. During training, the diffusion models are trained to reconstruct the GT given the noisy version of GT with a randomly sampled noise strength. However, during inference, the diffusion models are used to predict the clean data given output at the previous timestep. This would result in a train-test distribution gap for the input of the diffusion model, which can be problematic. Basically, this is because the denoiser is only trained to remove the noise and refine the mask. Thus, when the hidden mask is visible to the naked eye from the noisy input, the denoiser would simply remove the noise even if the hidden mask is incorrect.  As shown in \figref{fig:noise-target}, Noise-on-GT sharpens the first prediction $x_0^T$ but cannot correct it during the test.

There are many options to reduce the distribution gap. One possible choice is adopting the initial predictions from the base segmentation models as noise targets, but it is sub-optimal. As shown in \figref{fig:noise-target}, we could observe that the final prediction is largely determined by the first prediction which is denoised from pure noisy mask sampled from $q(x_T)$. As a result, it is straightforward that the first/initial predictions from the diffusion model itself are better noise targets than those from the base segmentation model. If we take a step further, we could imagine that the ideal case is to unroll the denoising iterations to obtain the $x_0^t$ that the training process needs, starting from $x_T$. However, we could not practically unroll the entire denoising iterations during training as it would require unaffordable computation and memory costs. To this end, we opt for a special case of the unrolling strategy that only unrolls one step, which is named as Noise-on-First-Prediction.

\textbf{Replace-Only Transition.} Different discrete diffusion models ~\cite{austin2021structured, gu2022vector, chang2023muse} mostly differ by their data corruption schemes. Starting from the uniform transition, \ie, corrupting discrete data by randomly resampling the categories uniformly, further researches also explore the corruption process with an additional special [MASK] token. The trends are the increasing portion of transition on [MASK] and even pure mask-only transition, which shows stronger performance for image generation ~\cite{chang2023muse} and connections to mask-based generative models ~\cite{chang2022maskgit}. 
However, instead of following the trend, we find that the replace-only transition, \ie, uniform transition, simply works better for segmentation. This is reasonable as the information density of segmentation maps is much lower than images. The corrupting by masking scheme would give a stronger signal that the pixels are incorrect, which would result in insufficient training as the model can easily infer the masked pixels by their neighbors without knowing about the conditional features from the base segmentation model. Replace-only transition makes the denoising process harder, forcing the model to pay more attention to conditional features, instead of lazy copy-and-paste style denoising of input noisy mask. 
Moreover, perhaps surprisingly, we find that none of these transitions matters much despite the replace-only indeed working slightly better, as reflected by our experiments.

\textbf{Two-Stage Training.} As previously discussed, the final refined mask is strongly related to the first-step refinement. Therefore, we propose a two-stage training strategy to improve the quality of the first-step results. In the first stage, we train the model for a single diffusion step with maximum random noise, which corresponds to the first denoising step. In the second stage, we finetune the model with all diffusion steps. This strategy provides a better initialization for the first predictions that benefits the Noise-on-First-Prediction strategy to further reduce the train-test distribution gap.

\textbf{Training Criterion.} With the $x_0$-parameterization, traditional discrete diffusion models ~\cite{austin2021structured} usually employ a hybrid loss with a variational bound (VLB) loss for the log-likelihood and an auxiliary cross-entropy loss for $x_0$ predictions. In this work, we empirically find that training with cross-entropy loss only leads to improved performance. This might be caused by the instability of VLB loss which needs an extra diffusion step before computing the loss.

\subsection{Inference}

In the following, we introduce the techniques for the inference of diffusion segmentation models. Notably, we find that the diffusion (re-noising) process during the inference that exactly follows the math might not be that important, and different denoising strategies influence more on the overall performance rather than multi-step gains.

\textbf{Free Re-noising.} The inference process of diffusion models follows a denoising trajectory that starts from a pure noisy mask $x_T$ and gradually predicts a less noisy mask $x_{t-1}$ given the current mask $x_t$. Mathematically, the denoising process of a diffusion model aims at learning the reverse transition $p(x_{t-1} | x_t)$. Instead of learning to predict $x_{t-1}$ given $x_t$ directly, recent works ~\cite{ho2020denoising, austin2021structured} typically adopt a reformulation, \ie, $x_0$-parameterization, where they predict the final state $x_0$ and sample $x_{t-1}$ by marginalizing,
\[
p(x_{t-1} | x_t) = \mathbb{E}_{x_0} \left[ q(\bx_{t-1}|\bx_{t}, \bx_0)  \right].
\]
In a sense, this strategy predicts $x_{t-1}$ by re-noising the prediction $x_0$ in the company with the previous prediction $x_t$.
This works well for image generation, but our exploration suggests that re-noising without the constraints of previous prediction $x_t$ works better for semantic segmentation. Specifically, we directly use the $q(x_{t-1}|x_0)$ for re-noising the $x_0$ prediction for the next denoising iteration. This indicates that the strict formulation of denoising diffusion modeling as ~\cite{ho2020denoising,austin2021structured} might not be that significant.  

\textbf{Stride Sampling.} Our $x_0$-parameterization and free re-noising strategies make it straightforward and easy to implement stride sampling for accelerating the iterative denoising process. In detail, this can be achieved by sampling the mask following the trajectory of $x_T$, $x_{T-\Delta_t}$, $x_{T-2\Delta_t}$, $x_0$, where $\Delta_t$ denotes the time stride and we predict $x_{t-\Delta_t}$ with $q(x_{t-\Delta_t}|x_0)$ and denoiser estimation on $x_0$.

\textbf{Classifier-free Guidance.} Classifier-free guidance ~\cite{ho2022classifier} (CFG) is frequently used in conditional image generation for increasing the influence of condition, \eg, text prompt. We also employ CFG for the proposed DDPS. The key idea is a trade-off between the logits of conditional model  $\ell_c$ and unconditional model  $\ell_u$ as $\ell = (s+1)\ell_c - s \ell_u$, where $s$ denotes the guidance scale.

%% file: sections/experiment.tex
\section{Experiments}

\subsection{Settings}

\textbf{Implementation Details.} 
We implement DDPS architecture in PyTorch~\cite{paszke2017automatic}. 
The base segmentation models are pre-trained and frozen during the training, and we directly use the checkpoints provided in MMSegmentation~\cite{mmseg2020}. 
The replace-only transition adopts a linear noise scheduler~\cite{ho2020denoising} that uniformly increases the rate of being replaced. We also test the cosine scheduler~\cite{nichol2021improved}, but found no improvement. We use 20 training steps for small-size models and 100 for the others, and we use 20 inference steps with stride sampling if not otherwise specified. 
All our models predict at the $1/4$ scale and we use a simple resizing codec with bi-linear interpolation to upscale the results. 
As previously discussed, we adopt a multi-stage training strategy. By default, we first pre-train the model on the single-step stage for 80k iterations and then finetune the model on the multi-step stage for another 160k iterations. 
For each stage, we use an initial learning rate of $1.5 \times 10^{-4}$, which is halved every 20k iterations until reach the minimum $1 \times 10^{-6}$. 
Similar to previous diffusion models~\cite{ho2020denoising}, we adopt EMA with a decay rate of 0.99 and an update interval of 25 iterations. 
We use a batch size of 32 for ADE20K and 16 for Cityscapes with the default data augmentations in MMSegmentation, \eg, scale jittering and random flipping.

\textbf{Denoiser.} We follow the UNet~\cite{ronneberger2015u} denoiser architecture as DDPM~\cite{ho2020denoising}. It takes as input the concatenation of the image feature map from the base segmentator and embedding of the discrete noisy mask along the channel dimension, and predicts a refined clean mask. 
We only tune the base dimensions and channel multipliers to construct two denoisers with different sizes, \ie, UNet-S (11.34M) and UNet-M (41.34M), to match the size of the base segmentator. 
More sophisticated designs are possible, but we opt for simplicity as they are not the main concerns of this work.

\subsection{Main Results}

To our knowledge, DDPS is the first general framework that introduces diffusion model to semantic segmentation tasks on common segmentation datasets like ADE20K and Cityscapes, different from previous diffusion segmentation methods \cite{amit2021segdiff, wolleb2022diffusion, wu2023medsegdiff}. 
In order to yield a fair comparison with most semantic segmentation methods, we evaluate the proposed DDPS with two representative segmentation models of CNN base and transformer base, \ie, DeepLabV3+~\cite{deeplabv3plus2018} (denoted as DDPS-DL) and Segformer~\cite{xie2021segformer} (denoted as DDPS-SF), and image backbones in different sizes, \ie, MobileNetV2~\cite{sandler2018mobilenetv2}, ResNet~\cite{he2016deep}, and MiT~\cite{xie2021segformer}. 
To clearly demonstrate the performance in geometric constraints, we report boundary IoU~\cite{cheng2021boundary} which describes the boundary errors in the segmentation map as a supplementary metric to mean IoU for each method.

\textbf{ADE20K.} \tabref{ade} summarized the results for ADE20K. We could observe that our DDPS consistently refines the results of base segmentation models with different sizes. Specifically, DDPS obtains the highest gains for the first step refinement and steadily acquires further improvement with more refinement steps. Notably, DDPS-SF-B0 achieves a remarkable gain against Segformer-B0 with 3.8 total mIoU improvement. Moreover, we could observe a larger improvement of 5 points in boundary IoU, which reflects the DDPS prior modeling abilities for geometric constraints on boundaries.

\textbf{Cityscapes.} We also provide the results on Cityscapes. As shown in \tabref{city}, DDPS also consistently improves the baseline segmentation models. The effect of DDPS refinement for Cityscapes is not as large as it is on ADE20K, this might be caused by that Cityscapes is more structural and easy so the existing models might be already sufficient to model the mask prior of geometric and semantic constraints.

\subsection{Ablation Study}

\textbf{Diffusion Transition.} 
\tabref{diffusion-transition} shows the results with different diffusion transition schemes. We compare our replace-only scheme with mask-only and hybrid replace-mask schemes. It can be seen that the performance difference is subtle while increasing the replacing rate benefits the first-step performance and our replace-only is slightly better for the final performance. Nevertheless, these results might indicate that the choice of diffusion transition is not the key to the success of diffusion segmentation.

\textbf{Diffusion Target.} 
 As previously discussed, existing diffusion that adds noise on ground truth for training would cause a distribution gap during the test. To verify it, we compare three different noise targets as shown in \tabref{noise-target}. We could observe that adding noise on ground truth yields obviously lower results than the others and the multi-step refinement makes the gap even higher (drop by over 10 mIoU). Adding noise on the initial prediction provided by the base segmentation model could achieve the same first-step performance as first prediction but is unable to obtain further gains. Instead, training with our first prediction scheme activates the progressive refinement with roughly 0.4 mIoU. This is because first predictions are better approximations for the test-time predictions than initial predictions.

\textbf{Inference Strategies.} During the test time, DDPS progressively refine the prediction $x_0$ by denoising a re-noised version of the previous prediction of $x_0$. In \tabref{inference-strategy}, we compare the conventional re-renoising strategy with posterior $q(x_{t-1}|x_{t},x_0)$ and our free re-noising strategy with $q(x_{t-1}|x_0)$. We could observe that our FRN simply works better, which indicates that the strict denoising process that follows the math might not be that important. Besides, it is also verified that the discrete CFG also brings further improvement.

\input{tables/ade}
\input{tables/city_all}

\input{tables/ablation}

\textbf{Training Strategies.} \tabref{train-strategy} evaluates the proposed training strategies with cross-entropy loss and multi-stage training. It can be seen both strategies mostly play the role of boosting the first-step performance without effect on the multi-step gains. Notably, the cross entropy loss significantly influences the performance, which might be explained by the instabilities of ELBO loss~\cite{ho2020denoising}.

\textbf{Iterative Refinement.} 
\tabref{inf-steps} shows the improvement brought by the iterative diffusion process of DDPS-DL and DDPS-SF in ADE20K dataset. After the diffusion mask prior introduced to the model, we could observe a significant improvement in the first step of the diffusion process obtains, which lifts the initial result by 0.5 and 1.8 in mIoU. With 20 steps of diffusion refinement, the models obtain an improvement of 0.1 and 0.8 in mIoU of DDPS-DL and DDPS-SF respectively. Along with the slight increase of mIoU, the quality of the segmentation mask improves observably which can partly be observed in the larger improvement of BIoU.

\subsection{Discussion}

\begin{figure}[t]
\centering
\includegraphics[width=1.0\linewidth]{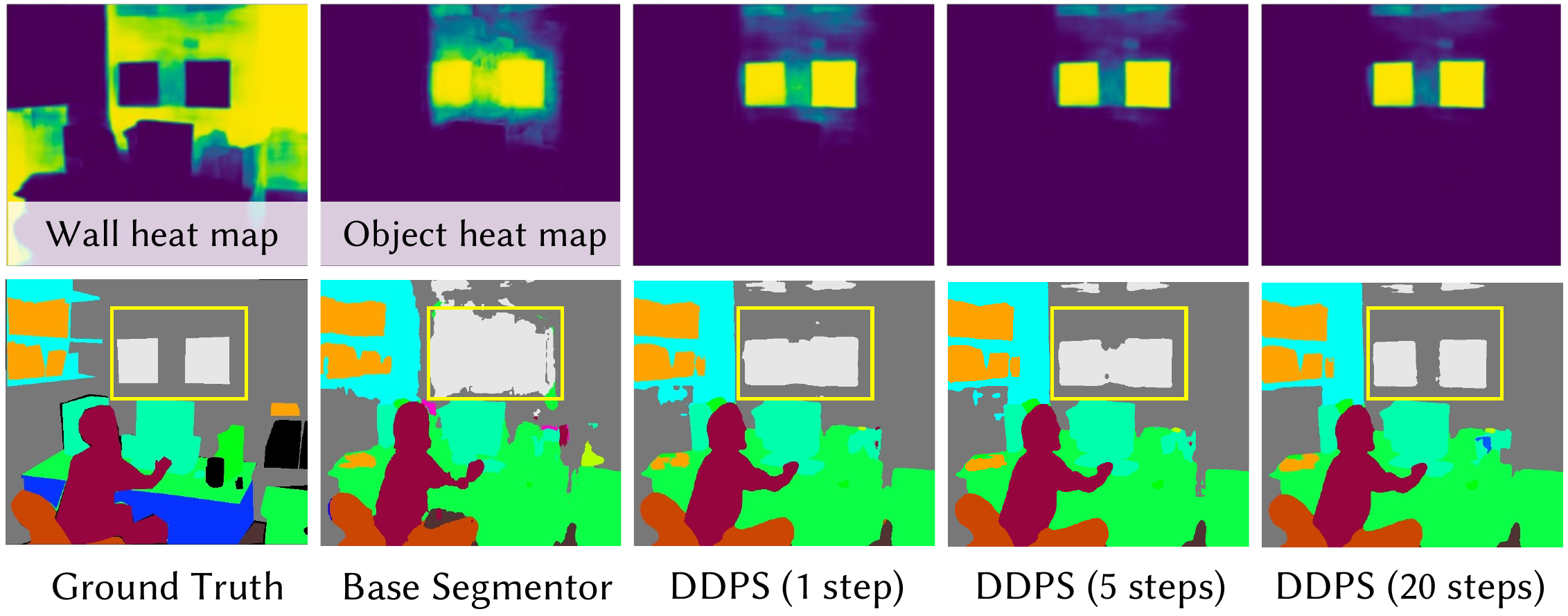}
\caption{The initial logits of base segmentor, in fact, roughly outline the objects, but the ambiguity of them between background (wall) leads to incorrect predictions. 
DDPS refinement work as a delineation of the logits based on prior knowledge about the objects.
}
\vspace{-10pt}
\label{fig:logits}
\end{figure}

\textbf{How many steps do we need ?} 
\tabref{steps} shows the results with different training and inference steps. Stride sampling is used when the training and inference steps are not matched. We could observe that increasing both training and inference steps benefit performance. Notably, it can be seen that 2-step stride sampling is already sufficient to reach the performance near the final one.

\begin{figure}[t]
\centering
\includegraphics[width=0.9\linewidth]{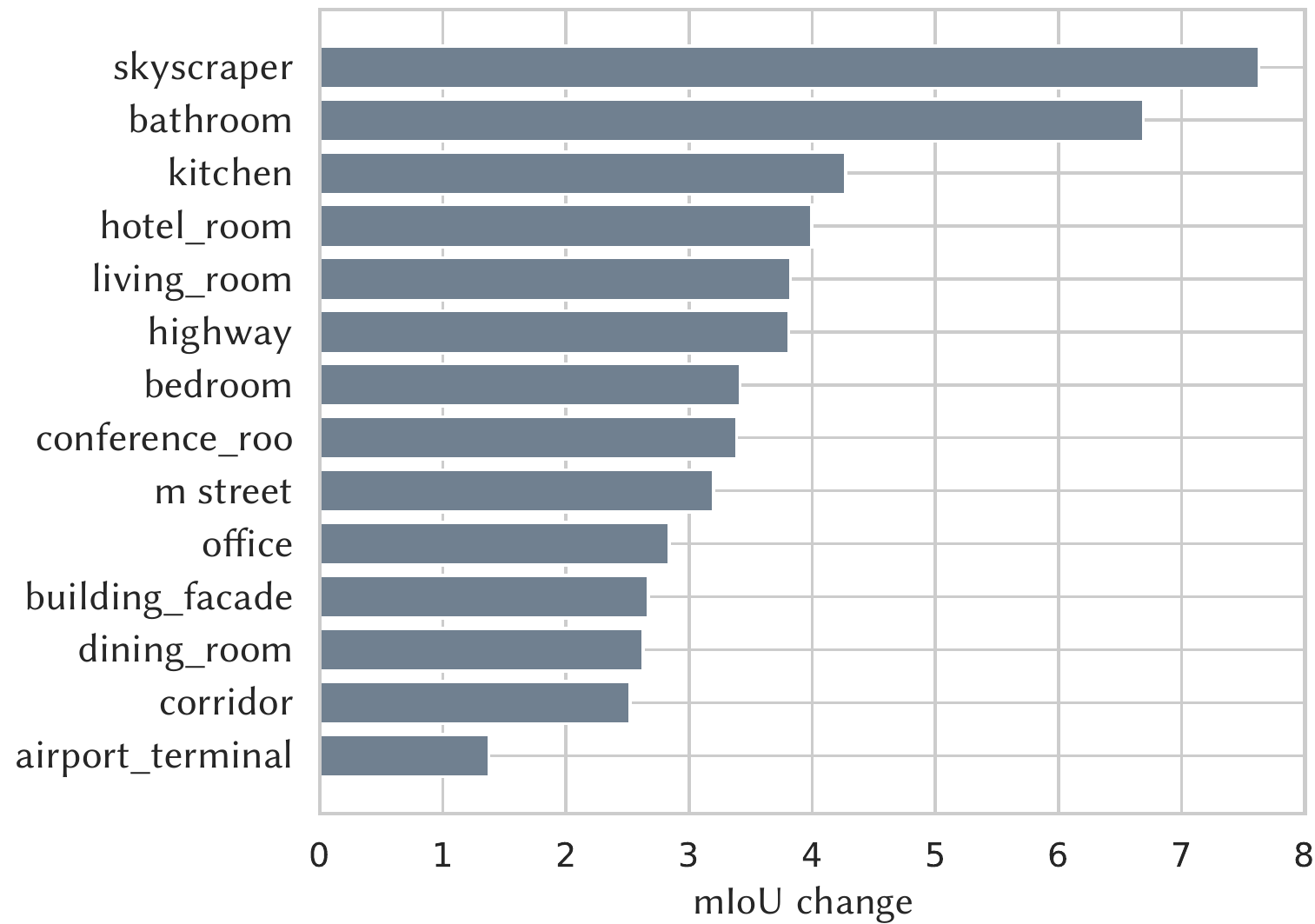}
\caption{The mIoU changes of 15 most common scenes. }
\vspace{-10pt}
\label{fig:stat}
\end{figure}

\textbf{What happens behind denoising diffusion ?} 
To analyze the question, we visualize the initial logits of the base segmentor and the refined logits of DDPS. As shown in \figref{fig:logits}, we could observe that the base segmentor actually generates roughly correct logits which highlight the objects. It is the ambiguity between objects and walls making the logits of walls higher than objects that results in errors. However, it is obvious that only highlight regions correspond to objects from the human side since we know no segmentation masks would look like that. DDPS works with a similar idea by imposing a mask prior that is modeled by the diffusion model, which can be reflected again from \figref{fig:logits}. For the first step of refinement, DDPS first delineates the logits based on the prior that masks are typically regular and sharp. Then it gradually lowers down the logits between two rectangles as it is more likely they are separate.  

\textbf{What situation does DDPS work better ?}
\figref{fig:stat} shows the mIoU change of 15 most common scenes in ADE20K. Considering the scenes with the same category, we can be observed that skyscraper achieves higher mIoU change than street and bathroom achieves higher change than other rooms. Intuitively, this is reasonable since these scenes contain more regular objects with limited categories where mask priors, \eg, geometric constraints, are more common.

%% file: tables/ade.tex
\begin{table}[t]
\small
\setlength{\tabcolsep}{0.05cm}
\centering
\begin{tabular}{l|c|c|c|c}
\hline
\textbf{Method}     & \textbf{Backbone} & \textbf{\#P(M)} & \textbf{mIoU(ss/ms)}  & \textbf{mBIoU} \\ 
\hline
PSPNet~\cite{zhao2017pyramid}       & MBNetV2   & 13.7  &  29.70/  -   & 16.47     \\
DeepLabV3+~\cite{deeplabv3plus2018} & MBNetV2   & 15.4  &  34.02*/  -   & 19.66     \\
OCRNet~\cite{yuan2020object}       & HRNetV2-18& 12.2  &  39.32*/40.80* & 24.16     \\
Segformer~\cite{xie2021segformer}   & MiT-B0    & 3.8   &  37.85*/38.97* & 21.73     \\
\rowcolor{graycolor}
DDPS-DL (ours)        & MBNetV2   & 27.9  &  38.07/39.07  &  23.47     \\
\rowcolor{graycolor}
DDPS-SF (ours)       & MiT-B0    & 15.1  &  41.67/41.95  &  26.77      \\ 
\hline
DeepLabV3+~\cite{deeplabv3plus2018} & ResNet50   & 43.7  & 43.95*/44.93*   & 19.66     \\
OCRNet~\cite{yuan2020object}       & HRNetV2-48 & 70.5  & 43.25*/44.88*   & 27.93    \\
Upernet~\cite{xiao2018unified}       & Swin-S     & 81.3  & 47.64/49.47   & 31.54    \\
ALTGVT-B~\cite{chu2021Twins}       & SVT-B      & 88.5  & 47.40/48.90   & 31.55    \\
Segformer~\cite{xie2021segformer}   & MiT-B2     & 27.5  & 46.80*/48.12*   & 29.25     \\
Segformer~\cite{xie2021segformer}  & MiT-B4     & 64.1  & 49.09*/50.72*   & 31.76      \\
\rowcolor{graycolor}
DDPS-DL (ours)       & ResNet50   & 54.2  & 45.26/46.01    &  30.39  \\
\rowcolor{graycolor}
DDPS-SF (ours)       & MiT-B2     & 65.5  & 49.73/49.96    &  33.95      \\ 
\hline
DeepLabV3+~\cite{deeplabv3plus2018} & ResNet101  & 62.7  & 45.47*/46.35*   & 19.66     \\
Upernet~\cite{xiao2018unified}       & Swin-B     & 121.4 & 48.13/49.72   & 30.44     \\
ALTGVT-L~\cite{chu2021Twins}       & SVT-L      & 133.0 & 48.80/50.20   & 32.49      \\
Segformer~\cite{xie2021segformer}   & MiT-B5     & 84.7  & 49.13*/50.22*   & 32.32      \\
\rowcolor{graycolor}
DDPS-DL (ours)       & ResNet101  & 104.3 & 46.32/46.99   & 31.82      \\
\rowcolor{graycolor}
DDPS-SF (ours)       & MiT-B5     & 122.8 & 51.11/51.71   & 35.66      \\ 
\hline
\end{tabular}
\caption{Comparison results on the ADE20K validation dataset. Our DDPS model surpasses the baseline and achieves significant improvement in  both mIoU and boundary IoU. ``\#P'' denotes the parameters of each model. ``MBNetV2'' denotes the MobileNet V2~\cite{sandler2018mobilenetv2},  ``{*}'' denotes the results reported are provided in MMSegmentation~\cite{mmseg2020}. ``-DL'' and ``-SF'' represent the base segmentation model are DeepLabV3+~\cite{deeplabv3plus2018} and Segformer~\cite{xie2021segformer}, respectively.}
\vspace{-10pt}
\label{tab:ade}
\end{table}

%% file: tables/city_all.tex
\begin{table}[t]
\small
\setlength{\tabcolsep}{0.05cm}
\centering
\begin{tabular}{l|c|c|c|c}
\hline
\textbf{Method}     & \textbf{Backbone} & \textbf{\#P(M)} & \textbf{mIoU(ss/ms)}  & \textbf{mBIoU} \\ 
\hline
PSPNet \cite{zhao2017pyramid}       & MBNetV2   & 13.7  &  70.23*/-      &  54.66     \\
DeepLabV3+ \cite{deeplabv3plus2018} & MBNetV2   & 15.4  &  75.20*/-       &  58.50     \\
OCRNet~\cite{yuan2020object}        & HRNetV2-18& 12.1  &  79.47*/80.91*  &  63.5    \\ 
Segformer \cite{xie2021segformer}   & MiT-B0    & 3.8   &  76.54*/78.22*  &  61.82      \\
\rowcolor{graycolor}
DDPS-DL (ours)                  & MBNetV2  &  27.9  &  78.11/79.73    &  61.14     \\ 
\rowcolor{graycolor}
DDPS-SF (ours)                  & MiT-B0  &  15.1  &  78.19/79.62    &  64.07      \\ 
\hline
DeepLabV3+ \cite{deeplabv3plus2018} & ResNet50     &  43.7   &  80.09*/81.13*  & 64.49 \\
OCRNet \cite{yuan2020object}        & HRNetV2-48   &  70.4   &  81.35*/82.70*  & 65.36  \\
Segformer \cite{xie2021segformer}   & MiT-B2       &  27.5   &  81.08*/82.18*  & 66.43  \\
\rowcolor{graycolor}
DDPS-DL (ours)                     & ResNet50    &  54.2   &  81.39/82.27   &  65.76  \\ 
\rowcolor{graycolor}
DDPS-SF (ours)                     & MiT-B2      &  65.5   &  81.77/82.22   &  67.75 \\ 
\hline
Segformer \cite{xie2021segformer}    & MiT-B5       & 84.7    &  82.25*/83.48* &  65.44  \\
DeepLabV3+ \cite{deeplabv3plus2018}  & ResNet101    & 62.7    &  80.97*/82.03* &  65.70  \\
\rowcolor{graycolor}
DDPS-DL (ours)                     & ResNet101    &  73.2   &  81.68/82.51   &  65.96  \\ 
\rowcolor{graycolor}
DDPS-SF (ours)                      & MiT-B5      &  122.8  &  82.42/82.94   &  66.45   \\
\hline
\end{tabular}
\caption{Comparison on the Cityscapes validation dataset.
}
\vspace{-10pt}
\label{tab:city}
\end{table}

%% file: tables/ablation.tex
\begin{table*}[ht]
   \begin{subtable}[h]{0.28\textwidth}
      \setlength{\tabcolsep}{0.1cm}
      \renewcommand{\arraystretch}{0.85}
       \centering
       \small
       \begin{tabular}{l|c|c}
    \toprule
      Transition & \one  & \twenty
      \\
       \midrule
       Mask Only  &  48.67  & 49.25 \\
       Replace+Mask  &  48.92 & 49.23 \\
       Replace Only &  48.88 & 49.30 \\
       \bottomrule
      \end{tabular}
      \vspace{-1mm}
      \caption{Diffusion transition.}
      \label{tab:diffusion-transition}
   \end{subtable}
   \hfill
   \begin{subtable}[h]{0.27\textwidth}
      \setlength{\tabcolsep}{0.1cm}
      \renewcommand{\arraystretch}{0.85}
       \centering
       \small
       \begin{tabular}{l|c|c}
        \toprule
      Diffusion on  & \one  & \twenty
      \\
       \midrule
        Ground Truth & 45.56 &  35.10 \\
        Init Prediction & 48.83 & 48.84  \\
       First Prediction  & 48.88 & 49.30  \\
       \bottomrule
      \end{tabular}
      \vspace{-1mm}
      \caption{Diffusion target.}
      \label{tab:noise-target}
   \end{subtable}
   \hfill
   \begin{subtable}[h]{0.19\textwidth}
      \setlength{\tabcolsep}{0.1cm}
      \renewcommand{\arraystretch}{0.85}
       \centering
       \small
       \begin{tabular}{c|c|c}
        \toprule
      FRN & CFG  & \twenty \\
       \midrule
       -   &  -  &  49.29 \\
       \checkmark  & - &  49.30 \\
       \checkmark  & \checkmark & 49.42  \\
      \bottomrule
      \end{tabular}
      \vspace{-1mm}
      \caption{Inference strategies.}
      \label{tab:inference-strategy}
   \end{subtable}
   \hfill
    \begin{subtable}[h]{0.22\textwidth}
      \setlength{\tabcolsep}{0.1cm}
      \renewcommand{\arraystretch}{0.85}
       \centering
       \small
       \begin{tabular}{c|c|c|c}
        \toprule
      CE  & MS  & \one & \twenty
      \\
       \midrule
       -  & \checkmark  & 45.18 & 45.67  \\
       \checkmark  & -  & 48.74 & 49.04  \\
       \checkmark  & \checkmark & 48.88 & 49.30 \\
      \bottomrule
      \end{tabular}
      \vspace{-1mm}
      \caption{Train strategies.}
      \label{tab:train-strategy}
   \end{subtable}
    \\
    \begin{subtable}[h]{0.51\textwidth}
   \setlength{\tabcolsep}{0.17cm}
   \renewcommand{\arraystretch}{0.85}
       \centering
       \small
       \begin{tabular}{l|c|c|c|c|c}
        \toprule
      Train / Test  & 2 steps & 5 steps & 20 steps & 50 steps & 100 steps
      \\
       \midrule
        20 steps & 49.40 & 49.45  & 49.44  & - &  - \\
        50 steps & 49.59  & 49.65 & 49.66 & 49.68 & - \\
       100 steps  & 49.69 & 49.71 &  49.73 & 49.74 & 49.75 \\
       \bottomrule
      \end{tabular}
      \vspace{-1mm}
      \caption{Effect of total train and test/inference steps.}
      \label{tab:steps}
    \end{subtable}
    \begin{subtable}[h]{0.49\textwidth}
   \setlength{\tabcolsep}{0.17cm}
   \renewcommand{\arraystretch}{0.7}
       \centering
       \small
       \begin{tabular}{l|c|c}
        \toprule
      Method  & DDPS-DL(mIoU/BIoU) & DDPS-SF(mIoU/BIoU)
      \\
       \midrule
       Initial & 37.37/22.63  &  39.02 / 23.82 \\
       \one    & 37.90/23.28  &  40.89 / 25.75 \\
       \twenty  & 38.07/23.47  & 41.67 / 26.77 \\
       \bottomrule
      \end{tabular}
      \vspace{-1mm}
      \caption{Effect of iterative refinement.}
      \label{tab:inf-steps}
    \end{subtable}
    \vspace{-2mm}
    \caption{\textbf{Ablation studies on ADE20K}. To reduce the computation cost, all the experiments are conducted on DDPS-SF with MiT-B2 backbone and UNet-S denoiser except the model in \tabref{inf-steps} use MiT-B0 and MobileNet V2 backbone. ``FRN'' denotes free re-noising strategy and ``CFG'' denotes classsifier-free guidance. ``CE'' denotes use cross entropy loss in training process, ``MS'' denotes multi-stage training. 
    By default, ``CFG'' is not used and the training steps are 100.}
    \label{tab:ablation}
    \vspace{-10pt}
\end{table*}

%% file: sections/conclusion.tex
\section{Conclusion}

In this work, we build up a new paradigm for semantic segmentation with a generative framework that explores the denoising diffusion model for the prior modeling of segmentation tasks.
Our explorations suggest several key points for the success of the application, \eg, the proper choice of noise target, training criterion, and inference strategies. 
Extensive experiments demonstrate the effectiveness of our models.
We hope this work could inspire further research on the prior modeling of task outcomes and diffusion models for visual perception.

%% file: sections/appendix.tex
\appendix

\section{Background}

Denoising Diffusion Probability Models (DDPM) \cite{ho2020denoising} aim at learning the target data distribution $p(x)$ and drawing samples from the distribution. Typically, diffusion models are used for modeling the distribution of media contents, \eg, images \cite{rombach2022high} and videos \cite{ho2022video}, but this work focuses on utilizing the diffusion models to model the distribution of segmentation mask. 

Basically, diffusion models generate samples of a data distribution $p(x)$ by learning the transformation from a known stationary distribution $q(x_T)$, \eg, standard Gaussian, to the target distribution. Instead of learning direct mapping, diffusion models learn a denoiser that takes a series of small denoising steps to gradually approach the target distribution. To achieve it, diffusion models first define a forward Markov chain that progressively diffuses (adds noise to) the target distribution into the stationary one with a transition distribution $q(x_{t}|x_{t-1})$. Then, the denoiser is trained to reverse the known forward process by learning the reverse transition $p(x_{t-1}|x_t)$, which is typically approximated by $q(x_{t-1}|x_t,x_0)$ and a denoiser that predicts $\hat{x}_0$. 

In the following, we include the basic derivations of three common denoising diffusion models, \ie, discrete diffusion \cite{austin2021structured,gu2022vector}, Gaussian diffusion \cite{ho2020denoising}, and bit diffusion \cite{chen2022analog}. We refer interested readers to the original papers for more details.

\textbf{Discrete Diffusion.}
For example, the discrete diffusion applies to discrete data with a fixed number of categories and takes the form of forward transition as, 
\begin{equation}
 q(\bx_t|\bx_{t-1}) = \mathrm{Cat}(\bx_t;\bs p = \bx_{t-1}\bs Q_t),
\end{equation}
where $Q_t$ defines the transition probabilities matrix for each category. Notably, it is not necessary to repeatly apply $q(x_t|x_{t-1})$ to obtain noisy $x_t$ from $x_0$. Instead, we could derive $q(x_t|x_0)$ analytically as,
\begin{equation}
\begin{aligned}
q\left(\boldsymbol{x}_t \mid \boldsymbol{x}_0\right)=\operatorname{Cat}\left(\boldsymbol{x}_t ; \boldsymbol{p}=\boldsymbol{x}_0 \overline{\boldsymbol{Q}}_t\right), \\
\quad \text { with } \quad \overline{\boldsymbol{Q}}_t=\boldsymbol{Q}_1 \boldsymbol{Q}_2 \ldots \boldsymbol{Q}_t
\end{aligned}
\end{equation}

Starting from a sample of the stationary distribution $x_T$, the reversing process iteratively applies the denoiser to denoise $x_t$ into $x_0$ and derive $x_{t-1}$ with the marginal distribution as,
\begin{equation}
\begin{aligned}
    q(\bx_{t-1}|\bx_{t}, \bx_0) &= \frac{q(\bx_{t}|\bx_{t-1}, \bx_0)q(\bx_{t-1}|\bx_0)}{q(\bx_{t}| \bx_0)} \\
    &=\mathrm{Cat}\left(\bx_{t-1};\bs p= \frac{\bx_t\bs Q_t^{\top} \odot  \bx_0 \overline{\bs Q}_{t-1}  }{\bx_0 \overline{\bs Q}_{t} \bx_t^\top}\right),
\end{aligned}
\end{equation}
forming a reverse trajectory, \ie, $x_T \rightarrow x_{T-1} \rightarrow \cdots \rightarrow, x_0$. 

To train the denoiser that models the $p_\theta(x_0|x_t; y)$, the typical practice employs an objective that minimizes the variational lower bound \cite{sohl2015deep} as
\begin{equation}
\begin{aligned}
\mathcal{L}_{\mathrm{vlb}} & =\mathcal{L}_0+\mathcal{L}_1+\cdots+\mathcal{L}_{T-1}+\mathcal{L}_T \\
\mathcal{L}_0 & =-\log p_\theta\left(\boldsymbol{x}_0 \mid \boldsymbol{x}_1, \boldsymbol{y}\right), \\
\mathcal{L}_{t-1} & =D_{K L}\left(q\left(\boldsymbol{x}_{t-1} \mid \boldsymbol{x}_t, \boldsymbol{x}_0\right) \| p_\theta\left(\boldsymbol{x}_{t-1} \mid \boldsymbol{x}_t, \boldsymbol{y}\right)\right), \\
\mathcal{L}_T & =D_{K L}\left(q\left(\boldsymbol{x}_T \mid \boldsymbol{x}_0\right) \| p\left(\boldsymbol{x}_T\right)\right),
\end{aligned}
\end{equation}
with an auxiliary entropy \cite{gu2022vector, austin2021structured} loss that encourages the high fidelity of the prediction of noiseless data as,
\begin{equation}
\mathcal{L}_{x_0}=-\log p_\theta\left(\boldsymbol{x}_0 \mid \boldsymbol{x}_t, \boldsymbol{y}\right).
\end{equation}

\textbf{Gaussian Diffusion.} Similarly, the Gaussian diffusion models \cite{ho2020denoising} define a Markovian noising process $q$ that gradually add noise to the data distribution $q(x_0)$ to generate a seqeuence of noisy data from $x_1$ to $x_T$. Specifically, the noising process is controlled by the forward transition defined by some variance schedule $\beta_t$ as,
\begin{alignat}{2}
q(x_t|x_{t-1}) &\coloneqq \mathcal{N}(x_t; \sqrt{1-\beta_t}x_{t-1}, \beta_t \mathbf{I}).
\end{alignat}
The direct forward transition $q(x_t|x_0)$  from $x_0$ to $x_t$ could also be derived as a Gaussian distribution as,
\begin{alignat}{2}
    q(x_t|x_0) &= \mathcal{N}(x_t; \sqrt{\bar{\alpha}_t} x_0, (1-\bar{\alpha}_t) \mathbf{I}) \\
    &= \sqrt{\bar{\alpha}_t} x_0 + \epsilon \sqrt{1-\bar{\alpha}_t},\text{  } \epsilon \sim \mathcal{N}(0, \mathbf{I}), \label{eq:jumpnoise}
\end{alignat}
where $\alpha_t \coloneqq 1 - \beta_t$ and $\bar{\alpha}_t \coloneqq \prod_{s=0}^{t} \alpha_s$. Besides, the posterior $q(x_{t-1}|x_t,x_0)$ is also a Gaussian with mean $\tilde{\mu}_t(x_t,x_0)$ and variance $\tilde{\beta}_t$ defined as follows:
\begin{alignat}{2}
    \tilde{\mu}_t(x_t,x_0) &\coloneqq \frac{\sqrt{\bar{\alpha}_{t-1}}\beta_t}{1-\bar{\alpha}_t}x_0 + \frac{\sqrt{\alpha_t}(1-\bar{\alpha}_{t-1})}{1-\bar{\alpha}_t} x_t \label{eq:mutilde} \\
    \tilde{\beta}_t &\coloneqq \frac{1-\bar{\alpha}_{t-1}}{1-\bar{\alpha}_t} \beta_t \label{eq:betatilde} \\
    q(x_{t-1}|x_t,x_0) &= \mathcal{N}(x_{t-1}; \tilde{\mu}(x_t, x_0), \tilde{\beta}_t \mathbf{I}) \label{eq:posterior}
\end{alignat}

To reverse the forward transition, one starts by sampling from $q(x_T)$ and sampling reverse steps $q(x_{t-1}|x_t)$ until we reach $x_0$. The denoiser is trained to approximate the $q(x_{t-1}|x_t)$, which is shown to be a diagonal Gaussian distribution as,
\begin{alignat}{2}
p_{\theta}(x_{t-1}|x_t) &\coloneqq \mathcal{N}(x_{t-1};\mu_{\theta}(x_t, t), \Sigma_{\theta}(x_t, t)) \label{eq:ptheta}
\end{alignat}

The typical training objective is the same as it is in discrete diffusion, which is the variational lower-bound $L_{\text{vlb}}$ for $p_{\theta}(x_0)$,
\begin{alignat}{2}
    L_{\text{vlb}} &\coloneqq L_0 + L_1 + ... + L_{T-1} + L_T \label{eq:loss} \\
    L_{0} &\coloneqq -\log p_{\theta}(x_0 | x_1) \label{eq:loss0} \\
    L_{t-1} &\coloneqq \kld{q(x_{t-1}|x_t,x_0)}{p_{\theta}(x_{t-1}|x_t)} \label{eq:losst} \\
    L_{T} &\coloneqq \kld{q(x_T | x_0)}{p(x_T)} \label{eq:lossT}
\end{alignat}

However, it is more common to use a different $x_0$-parameterization and objective for recent Gaussian diffusion models. In a nutshell, instead of training a denoiser to parameterize $\mu_{\theta}(x_t,t)$, they train a model $\epsilon_{\theta}(x_t,t)$ to predict $\epsilon$ from Equation \eqref{eq:jumpnoise}, and adopt a simplified objective that is defined as follows:
\begin{alignat}{2}
    L_{\text{simple}} &\coloneqq E_{t \sim [1,T],x_0 \sim q(x_0), \epsilon \sim \mathcal{N}(0, \mathbf{I})}[||\epsilon - \epsilon_{\theta}(x_t, t)||^2] \label{eq:lsimple}
\end{alignat}

\textbf{Bit Diffusion.} Bit diffusion \cite{chen2022analog} is an extension of Gaussian diffusion for discrete data. To achieve the application, Bit diffusion adopts a bit codec that encodes the discrete data into binary bits, \eg, a discrete variable with $K$ possible values can be encoded with $\log_2 K$ bits, as $\{0,1\}^n$. Bit diffusion directly treats binary bits as continuous variables, which they called analog bits, and applies the Gaussian diffusion model \cite{ho2020denoising} on them. The decoding is achieved by thresholding the analog bits and transforming bits back to the discrete value. In Pix2Seq-D \cite{chen2022generalist}, bit diffusion is applied to panoptic segmentation with improved techniques, \eg, input scaling (a way to increase the noise strength) and softmax cross entropy loss (compute loss based on the additional logit output), but it is not evaluated on semantic segmentation.

\section{More Results}

\tabref{ade-add} shows the additional results on ADE20K. It has been reported that training with $640\times 640$ crop size could improve the performance on ADE20K \cite{xie2021segformer}. Hence, we also experiment with this strategy on DDPS-SF and MiT-B5. We compare the results with its base segmentation model, \ie, Segformer-B5 trained on $640\times 640$. It can be observed that our DDPS-SF$^\dagger$ also improves the performance of the base segmentation model for both single- and multi-scale testing. Notably, DDPS-SF$^\dagger$ showed a 2-point improvement in boundary IoU, further demonstrating the effectiveness of DDPS for mask prior modeling. Additionally, we evaluated DDPS with a larger base segmentation model, i.e., Vit-Adapter-L \cite{chen2022vitadapter} with UperNet \cite{xiao2018unified}.
While the performance of Vit-Adapter-L-UperNet is already high, it has been observed that incorporating mask prior modeling using DDPS still yields benefits in this scenario.

\begin{table}[t]
\small
\setlength{\tabcolsep}{0.05cm}
\renewcommand{\arraystretch}{0.85}
\centering
\begin{tabular}{l|c|c|c|c}
\toprule
\textbf{Method}     & \textbf{Backbone} & \textbf{\#P(M)} & \textbf{mIoU(ss/ms)}  & \textbf{mBIoU} \\ 
\midrule
Segformer$^\dagger$~\cite{xie2021segformer}   & MiT-B5     & 84.7  & 51.13*/51.66*   & 34.34      \\
ViT-Adapter~\cite{chen2022vitadapter}& ViT-Adapter-L & 451.9 & 58.03/58.38 & 39.50 \\
\rowcolor{graycolor}
DDPS-SF$^\dagger$ (ours)       & MiT-B5     & 122.8 & 51.98/52.47   & 36.30      \\ 
\rowcolor{graycolor}
DDPS-UN (ours)       & ViT-Adapter-L & 617.3 & 58.14/58.48    & 39.69      \\
\bottomrule
\end{tabular}
\caption{Comparison results on the ADE20K validation dataset. Our DDPS model surpasses the baseline and achieves significant improvement in both mIoU and boundary IoU. ``\#P'' denotes the parameters of each model. ``{$\dagger$}'' denotes the methods are trained on $640\times 640$ image resolution. ``{*}'' denotes the results reported are provided in MMSegmentation~\cite{mmseg2020}. ``-SF'' and ``-UN'' represent the base segmentation model are Segformer~\cite{xie2021segformer} and UperNet~\cite{xiao2018unified}, respectively.}
\vspace{-10pt}
\label{tab:ade-add}
\end{table}

\section{More Implementation Details}

\textbf{Denoiser Archiecture.} As discussed in the main paper, we directly utilize the U-shape denoiser from DDPM \cite{ho2020denoising}. Here, we give a summary of the overall network structure of the adopted denoiser. Specifically, the denoiser follows an encoder-decoder architecture with skip connections that connect the different layers across the encoder and decoder at the same level. Each encoder/decoder layer is made up of two convolutional residual blocks and an attention block. The timestep is injected into each layer by scaling and shifting the intermediate results. 

\textbf{Training \& Inference Algorithms.} Algorithm \ref{alg:train} and \ref{alg:inference} detail the training and inference strategies of DDPS, which were discussed in the main paper.

\section{More Disscussion}

\textbf{Effect of Denoiser Size.} In our main paper, we constructed two denoisers, UNet-S (11.34M) and UNet-M (41.34M), with different sizes to match the base segmentator. In this section, we present an empirical study on the impact of denoiser size on performance. To further illustrate this impact, we constructed two additional denoisers, namely UNet-T (3.38M) and UNet-L (158.53M).
\tabref{denoiser-size} displays the results of DDPS-SF on ADE20K with denoisers of different sizes. Our observations suggest that there may be an upper and lower bound for denoiser size. A denoiser that is larger than the upper bound does not lead to improved performance, while a denoiser that is smaller than the lower bound may degrade performance. For example, decreasing the model size of DDPS-SF (MiT-B0) from UNet-S to UNet-T significantly degrades performance, while decreasing the model size of DDPS-SF (MiT-B5) from UNet-L to UNet-M does not affect performance. As a result, we recommend UNet-S for base segmentation models with less than 10M parameters and UNet-M for most cases (\>10M). UNet-L might be helpful for larger models (>500M), but UNets that are larger than UNet-L are unlikely to contribute more.

\input{tables/train_algo}

\textbf{Effect of EMA.} Exponential moving average (EMA) is a commonly used technique in the training of diffusion models \cite{ho2020denoising,austin2021structured}. In our DDPS, we also adopt the EMA and find that it has considerable importance for performance. \tabref{ema} displays the results of DDPS-SF with MiT-B2 backbone trained with and without EMA. We observe that EMA improves the performance of models trained on both 20 and 100 training steps. Specifically, we note that the impact of EMA is more pronounced for 100 training steps, where performance significantly degrades in the absence of EMA. This phenomenon may be attributed to larger instabilities when the model is trained for 100 steps.

\begin{table}[t]
\small
\setlength{\tabcolsep}{0.14cm}
\renewcommand{\arraystretch}{0.85}
\centering
\begin{tabular}{l|c|c|c|c|c}
\toprule
\textbf{Method}     & \textbf{Backbone} & \textbf{Denoiser} & \textbf{\#P(M)} & \textbf{mIoU}  & \textbf{mBIoU} \\ 
\midrule
DDPS-SF        & MiT-B0   & UNet-T   & 7.1 & 39.12  & 24.09      \\ 
\rowcolor{graycolor} 
DDPS-SF        & MiT-B0   & UNet-S   & 15.1 & 41.67   & 26.77      \\ 
\midrule
DDPS-SF       & MiT-B2   & UNet-S   & 36.0 & 49.33   & 33.20      \\ 
\rowcolor{graycolor} 
DDPS-SF      & MiT-B2   & UNet-M   & 65.5 & 49.73   & 33.95      \\ 
\midrule
\rowcolor{graycolor} 
DDPS-SF      & MiT-B5   & UNet-M   & 122.8 & 51.11   & 35.66      \\ 
DDPS-SF      & MiT-B5   & UNet-L   & 239.9 & 51.13   & 35.62      \\ 
\bottomrule
\end{tabular}
\caption{Effect of different denoiser size. }
\label{tab:denoiser-size}
\end{table}

\begin{table}[t]
\small
\setlength{\tabcolsep}{0.18cm}
\renewcommand{\arraystretch}{0.85}
\centering
\begin{tabular}{l|c|c|c|c|c}
\toprule
\textbf{Method}  & \textbf{backbone}  & \textbf{Steps}  & \textbf{EMA} & \textbf{mIoU}  & \textbf{mBIoU} \\ 
\midrule
DDPS-SF   & MiT-B2 & 20       & -   & 49.30   & 33.41      \\ 
DDPS-SF   & MiT-B2 & 20      & \checkmark   & 49.44   & 33.50      \\ 
DDPS-SF    & MiT-B2 & 100       & -    & 49.25   & 33.63      \\ 
\rowcolor{graycolor} 
DDPS-SF    & MiT-B2 &  100      & \checkmark  & 49.73   & 33.95      \\ 
\bottomrule
\end{tabular}
\caption{Effect of EMA. Steps refer to the total timesteps during the training.}
\label{tab:ema}
\end{table}

\begin{table}[t]
\small
\setlength{\tabcolsep}{0.19cm}
\renewcommand{\arraystretch}{0.85}
\centering
\begin{tabular}{l|c|c|c|c}
\toprule
\textbf{Method}  & \textbf{backbone} & \textbf{\#P(M)} &  \textbf{mIoU} & \textbf{mBIoU}\\ 
\midrule
Segformer-Unet   & MiT-B2           & 65.5            &  48.69          &  32.77      \\
\rowcolor{graycolor} 
DDPS-SF    & MiT-B2           & 65.5            &  49.73         &  33.95      \\ 
\bottomrule
\end{tabular}
\caption{Comparison results to Unet as additional layers of decode head.}
\label{tab:add_unet}
\end{table}

\input{tables/test_algo}

\begin{table}[t]
\small
\setlength{\tabcolsep}{0.14cm}
\renewcommand{\arraystretch}{0.85}
\centering
\begin{tabular}{l|c|c|c|c|c}
\toprule
\textbf{Method}     & \textbf{Backbone} & \textbf{Diffusion} & \textbf{\#P(M)} & \textbf{mIoU}  & \textbf{mBIoU} \\ 
\midrule
\rowcolor{graycolor}
DDPS-SF      & MiT-B2   & Discrete   & 65.5 & 49.73   & 33.95      \\ 
DDPS-SF        & MiT-B2   & Bit   & 65.5 & 48.98   & 33.20      \\ 
\bottomrule
\end{tabular}
\caption{Comparison results on different diffusion models.}
\label{tab:diff-diff}
\end{table}

\textbf{Effectiveness of Diffusion Denoising Modeling.}
Since our model introduces an additional UNet for the existing segmentation models, we further evaluate the models that treat UNet as additional layers of their decode head without diffusion denoising modeling. The comparison is shown in \tabref{add_unet}. We could observe that the diffusion mask prior improves the performance of mIoU for 1 point and BIoU for 1.2 points over models that simply treat Unet as layers of the decode head, which demonstrates the effectiveness of our diffusion modeling.

\textbf{Bit Instantiation of DDPS.} Here, we provide results of the instantiation of DDPS with bit codec and Gaussian diffusion, which is similar to Pix2Seq-D \cite{chen2022generalist} that applies diffusion models for panoptic segmentation.
Following \cite{chen2022generalist}, we adopt bit codec with the input scaling (scaling ratio = 0.1) and softmax cross entropy loss. \tabref{diff-diff} shows the results on ADE20K. We could observe that the instantiation of DDPS with discrete diffusion achieves better performance.

\section{More Visual Results}

\begin{figure*}[t]
\centering
\includegraphics[width=0.9\linewidth]{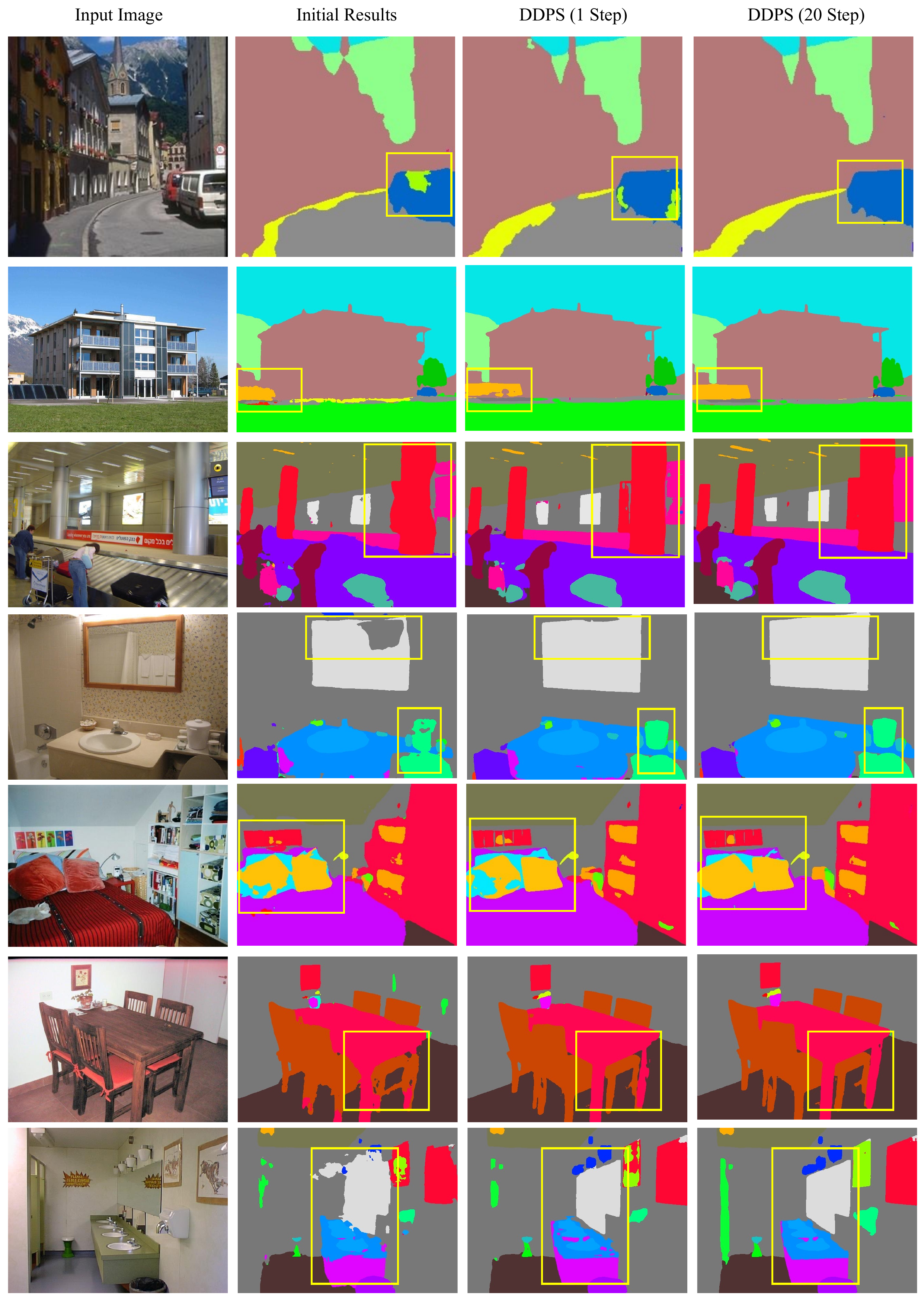}
\caption{Additional visual results on ADE20K. }
\label{fig:ade-more}
\end{figure*}

\textbf{ADE20K.} \figref{fig:ade-more} shows more visualization of the refinement trajectories of DDPS on ADE20K. We see that the model iteratively refines the initial results to obtain more smooth and regular results that better match the distribution of segmentation masks.

\begin{figure*}[t]
\centering
\includegraphics[width=1\linewidth]{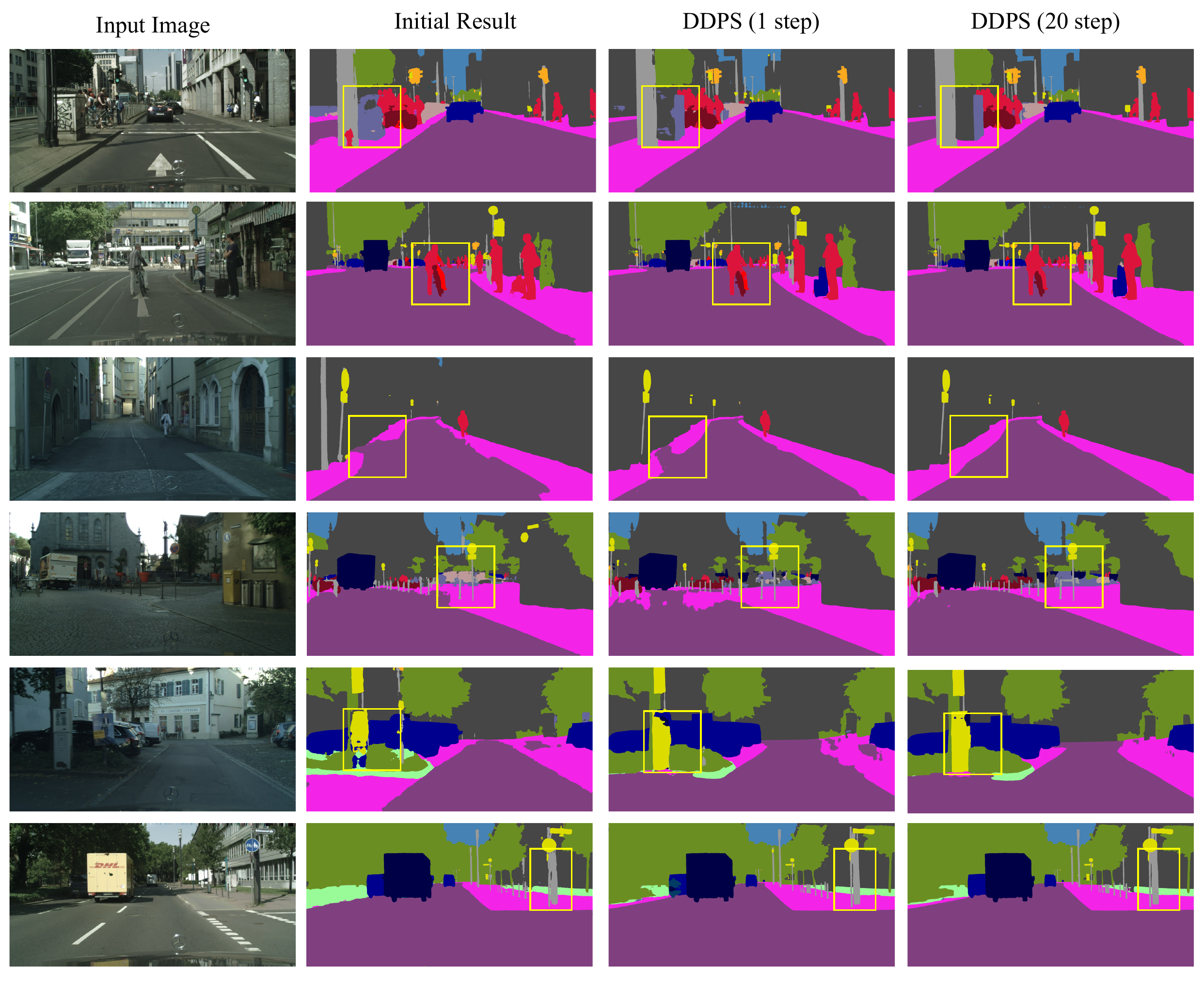}
\caption{Visual results on Cityscapes. }
\label{fig:city-more}
\end{figure*}

\textbf{Cityscapes.} \figref{fig:city-more} shows the visualization of the refinement trajectories of DDPS on Cityscapes.

%% file: tables/train_algo.tex
\begin{algorithm}[t]
\small
\caption{\small DDPS Training Algorithm.
}
\label{alg:train}
\definecolor{codeblue}{rgb}{0.25,0.5,0.5}
\definecolor{codekw}{rgb}{0.85, 0.18, 0.50}
\lstset{
  backgroundcolor=\color{white},
  basicstyle=\fontsize{7.5pt}{7.5pt}\ttfamily\selectfont,
  columns=fullflexible,
  breaklines=true,
  captionpos=b,
  commentstyle=\fontsize{7.5pt}{7.5pt}\color{codeblue},
  keywordstyle=\fontsize{7.5pt}{7.5pt}\color{codekw},
  escapechar={|}, 
}
\begin{lstlisting}[language=python]
def train_loss(images, masks):
  """images: [b, 3, h, w], masks: [b, h, w]."""
  
  # Get initial predictions in terms of features.
  h = base_segmentation_model(images)
  
  # Represent mask.
  m_latent = codec.encode(masks)

  # Get the first prediction.
  with no_grad():
    m_T = randint(0, num_classes)
    m_1 = denoiser(m_T, h, 0).argmax(1)

  # Add noise on the first prediction.
  t = uniform(0, train_steps)
  m_noisy = replace_only_transition(m_1, t)

  # Predict and compute loss.
  m_logits = denoiser(m_noisy, h, t)
  loss = cross_entropy(m_logits, masks)
  
  return loss.mean()
\end{lstlisting}
\end{algorithm}

%% file: tables/test_algo.tex
\begin{algorithm}[t]
\small
\caption{\small DDPS Inference Algorithm.}
\label{alg:inference}
\definecolor{codeblue}{rgb}{0.25,0.5,0.5}
\definecolor{codekw}{rgb}{0.85, 0.18, 0.50}
\lstset{
  backgroundcolor=\color{white},
  basicstyle=\fontsize{7.5pt}{7.5pt}\ttfamily\selectfont,
  columns=fullflexible,
  breaklines=true,
  captionpos=b,
  commentstyle=\fontsize{7.5pt}{7.5pt}\color{codeblue},
  keywordstyle=\fontsize{7.5pt}{7.5pt}\color{codekw},
  escapechar={|}, 
}
\begin{lstlisting}[language=python]
def inference(images, timesteps, cfg=1.0):
  """images: [b, h, w, 3]."""
  
  # Get initial predictions in terms of features.
  h = base_segmentation_model(images)

  m_t = randint()   # shape is [b,h,w].
  for t in range(timesteps):
  
    # Predict m_0 from m_t.
    m_pred = denoiser(m_t, h)

    # classifier free guidance
    if cfg != 1.0:
      m_pred_u = denoiser(m_t, None)
      m_pred = (cfg+1)*m_pred - cfg*m_pred_u
        
    # Estimate m at t_next.
    m_t = replace_only_transition(m_pred, t_next)
    
  # Decode into masks.
  masks = codec.decode(m_pred)
  return masks
\end{lstlisting}
\end{algorithm}

%% file: arxiv.bbl
\begin{thebibliography}{10}\itemsep=-1pt

\bibitem{amit2021segdiff}
Tomer Amit, Eliya Nachmani, Tal Shaharbany, and Lior Wolf.
\newblock Segdiff: Image segmentation with diffusion probabilistic models.
\newblock {\em arXiv preprint arXiv:2112.00390}, 2021.

\bibitem{austin2021structured}
Jacob Austin, Daniel~D Johnson, Jonathan Ho, Daniel Tarlow, and Rianne van~den Berg.
\newblock Structured denoising diffusion models in discrete state-spaces.
\newblock {\em Advances in Neural Information Processing Systems}, 34:17981--17993, 2021.

\bibitem{azuma1997survey}
Ronald~T Azuma.
\newblock A survey of augmented reality.
\newblock {\em Presence: teleoperators \& virtual environments}, 6(4):355--385, 1997.

\bibitem{bao2021beit}
Hangbo Bao, Li Dong, Songhao Piao, and Furu Wei.
\newblock Beit: Bert pre-training of image transformers.
\newblock {\em arXiv preprint arXiv:2106.08254}, 2021.

\bibitem{baranchuk2021label}
Dmitry Baranchuk, Ivan Rubachev, Andrey Voynov, Valentin Khrulkov, and Artem Babenko.
\newblock Label-efficient semantic segmentation with diffusion models.
\newblock {\em arXiv preprint arXiv:2112.03126}, 2021.

\bibitem{billinghurst2015survey}
Mark Billinghurst, Adrian Clark, Gun Lee, et~al.
\newblock A survey of augmented reality.
\newblock {\em Foundations and Trends{\textregistered} in Human--Computer Interaction}, 8(2-3):73--272, 2015.

\bibitem{brempong2022denoising}
Emmanuel~Asiedu Brempong, Simon Kornblith, Ting Chen, Niki Parmar, Matthias Minderer, and Mohammad Norouzi.
\newblock Denoising pretraining for semantic segmentation.
\newblock In {\em Proceedings of the IEEE/CVF Conference on Computer Vision and Pattern Recognition}, pages 4175--4186, 2022.

\bibitem{brown2020language}
Tom Brown, Benjamin Mann, Nick Ryder, Melanie Subbiah, Jared~D Kaplan, Prafulla Dhariwal, Arvind Neelakantan, Pranav Shyam, Girish Sastry, Amanda Askell, et~al.
\newblock Language models are few-shot learners.
\newblock {\em Advances in neural information processing systems}, 33:1877--1901, 2020.

\bibitem{burgert2022peekaboo}
Ryan Burgert, Kanchana Ranasinghe, Xiang Li, and Michael~S Ryoo.
\newblock Peekaboo: Text to image diffusion models are zero-shot segmentors.
\newblock {\em arXiv preprint arXiv:2211.13224}, 2022.

\bibitem{carion2020end}
Nicolas Carion, Francisco Massa, Gabriel Synnaeve, Nicolas Usunier, Alexander Kirillov, and Sergey Zagoruyko.
\newblock End-to-end object detection with transformers.
\newblock In {\em Computer Vision--ECCV 2020: 16th European Conference, Glasgow, UK, August 23--28, 2020, Proceedings, Part I 16}, pages 213--229. Springer, 2020.

\bibitem{chang2023muse}
Huiwen Chang, Han Zhang, Jarred Barber, AJ Maschinot, Jose Lezama, Lu Jiang, Ming-Hsuan Yang, Kevin Murphy, William~T Freeman, Michael Rubinstein, et~al.
\newblock Muse: Text-to-image generation via masked generative transformers.
\newblock {\em arXiv preprint arXiv:2301.00704}, 2023.

\bibitem{chang2022maskgit}
Huiwen Chang, Han Zhang, Lu Jiang, Ce Liu, and William~T Freeman.
\newblock Maskgit: Masked generative image transformer.
\newblock In {\em Proceedings of the IEEE/CVF Conference on Computer Vision and Pattern Recognition}, pages 11315--11325, 2022.

\bibitem{chen2014semantic}
Liang-Chieh Chen, George Papandreou, Iasonas Kokkinos, Kevin Murphy, and Alan~L Yuille.
\newblock Semantic image segmentation with deep convolutional nets and fully connected crfs.
\newblock {\em arXiv preprint arXiv:1412.7062}, 2014.

\bibitem{chen2017rethinking}
Liang-Chieh Chen, George Papandreou, Florian Schroff, and Hartwig Adam.
\newblock Rethinking atrous convolution for semantic image segmentation.
\newblock {\em arXiv preprint arXiv:1706.05587}, 2017.

\bibitem{deeplabv3plus2018}
Liang-Chieh Chen, Yukun Zhu, George Papandreou, Florian Schroff, and Hartwig Adam.
\newblock Encoder-decoder with atrous separable convolution for semantic image segmentation.
\newblock In {\em ECCV}, 2018.

\bibitem{chen2022diffusiondet}
Shoufa Chen, Peize Sun, Yibing Song, and Ping Luo.
\newblock Diffusiondet: Diffusion model for object detection.
\newblock {\em arXiv preprint arXiv:2211.09788}, 2022.

\bibitem{chen2022generalist}
Ting Chen, Lala Li, Saurabh Saxena, Geoffrey Hinton, and David~J Fleet.
\newblock A generalist framework for panoptic segmentation of images and videos.
\newblock {\em arXiv preprint arXiv:2210.06366}, 2022.

\bibitem{chen2022analog}
Ting Chen, Ruixiang Zhang, and Geoffrey Hinton.
\newblock Analog bits: Generating discrete data using diffusion models with self-conditioning.
\newblock {\em arXiv preprint arXiv:2208.04202}, 2022.

\bibitem{chen2022vitadapter}
Zhe Chen, Yuchen Duan, Wenhai Wang, Junjun He, Tong Lu, Jifeng Dai, and Yu Qiao.
\newblock Vision transformer adapter for dense predictions.
\newblock {\em arXiv preprint arXiv:2205.08534}, 2022.

\bibitem{cheng2021boundary}
Bowen Cheng, Ross Girshick, Piotr Doll{\'a}r, Alexander~C Berg, and Alexander Kirillov.
\newblock Boundary iou: Improving object-centric image segmentation evaluation.
\newblock In {\em Proceedings of the IEEE/CVF Conference on Computer Vision and Pattern Recognition}, pages 15334--15342, 2021.

\bibitem{cheng2022masked}
Bowen Cheng, Ishan Misra, Alexander~G Schwing, Alexander Kirillov, and Rohit Girdhar.
\newblock Masked-attention mask transformer for universal image segmentation.
\newblock In {\em Proceedings of the IEEE/CVF Conference on Computer Vision and Pattern Recognition}, pages 1290--1299, 2022.

\bibitem{cheng2021per}
Bowen Cheng, Alex Schwing, and Alexander Kirillov.
\newblock Per-pixel classification is not all you need for semantic segmentation.
\newblock {\em Advances in Neural Information Processing Systems}, 34:17864--17875, 2021.

\bibitem{chu2021Twins}
Xiangxiang Chu, Zhi Tian, Yuqing Wang, Bo Zhang, Haibing Ren, Xiaolin Wei, Huaxia Xia, and Chunhua Shen.
\newblock Twins: Revisiting the design of spatial attention in vision transformers.
\newblock In {\em NeurIPS 2021}, 2021.

\bibitem{mmseg2020}
MMSegmentation Contributors.
\newblock {MMSegmentation}: Openmmlab semantic segmentation toolbox and benchmark.
\newblock \url{https://github.com/open-mmlab/mmsegmentation}, 2020.

\bibitem{Cordts2016Cityscapes}
Marius Cordts, Mohamed Omran, Sebastian Ramos, Timo Rehfeld, Markus Enzweiler, Rodrigo Benenson, Uwe Franke, Stefan Roth, and Bernt Schiele.
\newblock The cityscapes dataset for semantic urban scene understanding.
\newblock In {\em Proc. of the IEEE Conference on Computer Vision and Pattern Recognition (CVPR)}, 2016.

\bibitem{daras2022soft}
Giannis Daras, Mauricio Delbracio, Hossein Talebi, Alexandros~G Dimakis, and Peyman Milanfar.
\newblock Soft diffusion: Score matching for general corruptions.
\newblock {\em arXiv preprint arXiv:2209.05442}, 2022.

\bibitem{deng2009imagenet}
Jia Deng, Wei Dong, Richard Socher, Li-Jia Li, Kai Li, and Li Fei-Fei.
\newblock Imagenet: A large-scale hierarchical image database.
\newblock In {\em 2009 IEEE conference on computer vision and pattern recognition}, pages 248--255. Ieee, 2009.

\bibitem{dieleman2022continuous}
Sander Dieleman, Laurent Sartran, Arman Roshannai, Nikolay Savinov, Yaroslav Ganin, Pierre~H Richemond, Arnaud Doucet, Robin Strudel, Chris Dyer, Conor Durkan, et~al.
\newblock Continuous diffusion for categorical data.
\newblock {\em arXiv preprint arXiv:2211.15089}, 2022.

\bibitem{dong2021polyp}
Bo Dong, Wenhai Wang, Deng-Ping Fan, Jinpeng Li, Huazhu Fu, and Ling Shao.
\newblock Polyp-pvt: Polyp segmentation with pyramid vision transformers.
\newblock {\em arXiv preprint arXiv:2108.06932}, 2021.

\bibitem{dosovitskiy2020image}
Alexey Dosovitskiy, Lucas Beyer, Alexander Kolesnikov, Dirk Weissenborn, Xiaohua Zhai, Thomas Unterthiner, Mostafa Dehghani, Matthias Minderer, Georg Heigold, Sylvain Gelly, et~al.
\newblock An image is worth 16x16 words: Transformers for image recognition at scale.
\newblock {\em arXiv preprint arXiv:2010.11929}, 2020.

\bibitem{fan2020pranet}
Deng-Ping Fan, Ge-Peng Ji, Tao Zhou, Geng Chen, Huazhu Fu, Jianbing Shen, and Ling Shao.
\newblock Pranet: Parallel reverse attention network for polyp segmentation.
\newblock In {\em Medical Image Computing and Computer Assisted Intervention--MICCAI 2020: 23rd International Conference, Lima, Peru, October 4--8, 2020, Proceedings, Part VI 23}, pages 263--273. Springer, 2020.

\bibitem{feng2020deep}
Di Feng, Christian Haase-Sch{\"u}tz, Lars Rosenbaum, Heinz Hertlein, Claudius Glaeser, Fabian Timm, Werner Wiesbeck, and Klaus Dietmayer.
\newblock Deep multi-modal object detection and semantic segmentation for autonomous driving: Datasets, methods, and challenges.
\newblock {\em IEEE Transactions on Intelligent Transportation Systems}, 22(3):1341--1360, 2020.

\bibitem{garcia2017review}
Alberto Garcia-Garcia, Sergio Orts-Escolano, Sergiu Oprea, Victor Villena-Martinez, and Jose Garcia-Rodriguez.
\newblock A review on deep learning techniques applied to semantic segmentation.
\newblock {\em arXiv preprint arXiv:1704.06857}, 2017.

\bibitem{gong2022diffuseq}
Shansan Gong, Mukai Li, Jiangtao Feng, Zhiyong Wu, and LingPeng Kong.
\newblock Diffuseq: Sequence to sequence text generation with diffusion models.
\newblock {\em arXiv preprint arXiv:2210.08933}, 2022.

\bibitem{goodfellow2020generative}
Ian Goodfellow, Jean Pouget-Abadie, Mehdi Mirza, Bing Xu, David Warde-Farley, Sherjil Ozair, Aaron Courville, and Yoshua Bengio.
\newblock Generative adversarial networks.
\newblock {\em Communications of the ACM}, 63(11):139--144, 2020.

\bibitem{gu2022vector}
Shuyang Gu, Dong Chen, Jianmin Bao, Fang Wen, Bo Zhang, Dongdong Chen, Lu Yuan, and Baining Guo.
\newblock Vector quantized diffusion model for text-to-image synthesis.
\newblock In {\em Proceedings of the IEEE/CVF Conference on Computer Vision and Pattern Recognition}, pages 10696--10706, 2022.

\bibitem{gu2022diffusioninst}
Zhangxuan Gu, Haoxing Chen, Zhuoer Xu, Jun Lan, Changhua Meng, and Weiqiang Wang.
\newblock Diffusioninst: Diffusion model for instance segmentation.
\newblock {\em arXiv preprint arXiv:2212.02773}, 2022.

\bibitem{he2022masked}
Kaiming He, Xinlei Chen, Saining Xie, Yanghao Li, Piotr Doll{\'a}r, and Ross Girshick.
\newblock Masked autoencoders are scalable vision learners.
\newblock In {\em Proceedings of the IEEE/CVF Conference on Computer Vision and Pattern Recognition}, pages 16000--16009, 2022.

\bibitem{he2017mask}
Kaiming He, Georgia Gkioxari, Piotr Doll{\'a}r, and Ross Girshick.
\newblock Mask r-cnn.
\newblock In {\em Proceedings of the IEEE international conference on computer vision}, pages 2961--2969, 2017.

\bibitem{he2016deep}
Kaiming He, Xiangyu Zhang, Shaoqing Ren, and Jian Sun.
\newblock Deep residual learning for image recognition.
\newblock In {\em Proceedings of the IEEE conference on computer vision and pattern recognition}, pages 770--778, 2016.

\bibitem{heide2016proximal}
Felix Heide, Steven Diamond, Matthias Nie{\ss}ner, Jonathan Ragan-Kelley, Wolfgang Heidrich, and Gordon Wetzstein.
\newblock Proximal: Efficient image optimization using proximal algorithms.
\newblock {\em ACM Transactions on Graphics (TOG)}, 35(4):1--15, 2016.

\bibitem{ho2020denoising}
Jonathan Ho, Ajay Jain, and Pieter Abbeel.
\newblock Denoising diffusion probabilistic models.
\newblock {\em Advances in Neural Information Processing Systems}, 33:6840--6851, 2020.

\bibitem{ho2022classifier}
Jonathan Ho and Tim Salimans.
\newblock Classifier-free diffusion guidance.
\newblock {\em arXiv preprint arXiv:2207.12598}, 2022.

\bibitem{ho2022video}
Jonathan Ho, Tim Salimans, Alexey Gritsenko, William Chan, Mohammad Norouzi, and David~J Fleet.
\newblock Video diffusion models.
\newblock {\em arXiv preprint arXiv:2204.03458}, 2022.

\bibitem{kadkhodaie2020solving}
Zahra Kadkhodaie and Eero~P Simoncelli.
\newblock Solving linear inverse problems using the prior implicit in a denoiser.
\newblock {\em arXiv preprint arXiv:2007.13640}, 2020.

\bibitem{kim2021unsupervised}
Dahye Kim and Byung-Woo Hong.
\newblock Unsupervised segmentation incorporating shape prior via generative adversarial networks.
\newblock In {\em Proceedings of the IEEE/CVF International Conference on Computer Vision}, pages 7324--7334, 2021.

\bibitem{kingma2013auto}
Diederik~P Kingma and Max Welling.
\newblock Auto-encoding variational bayes.
\newblock {\em arXiv preprint arXiv:1312.6114}, 2013.

\bibitem{kong2020diffwave}
Zhifeng Kong, Wei Ping, Jiaji Huang, Kexin Zhao, and Bryan Catanzaro.
\newblock Diffwave: A versatile diffusion model for audio synthesis.
\newblock {\em arXiv preprint arXiv:2009.09761}, 2020.

\bibitem{li2021semantic}
Daiqing Li, Junlin Yang, Karsten Kreis, Antonio Torralba, and Sanja Fidler.
\newblock Semantic segmentation with generative models: Semi-supervised learning and strong out-of-domain generalization.
\newblock In {\em Proceedings of the IEEE/CVF Conference on Computer Vision and Pattern Recognition}, pages 8300--8311, 2021.

\bibitem{li2022diffusion}
Xiang~Lisa Li, John Thickstun, Ishaan Gulrajani, Percy Liang, and Tatsunori~B Hashimoto.
\newblock Diffusion-lm improves controllable text generation.
\newblock {\em arXiv preprint arXiv:2205.14217}, 2022.

\bibitem{li2022bevformer}
Zhiqi Li, Wenhai Wang, Hongyang Li, Enze Xie, Chonghao Sima, Tong Lu, Yu Qiao, and Jifeng Dai.
\newblock Bevformer: Learning bird’s-eye-view representation from multi-camera images via spatiotemporal transformers.
\newblock In {\em Computer Vision--ECCV 2022: 17th European Conference, Tel Aviv, Israel, October 23--27, 2022, Proceedings, Part IX}, pages 1--18. Springer, 2022.

\bibitem{liu2021swin}
Ze Liu, Yutong Lin, Yue Cao, Han Hu, Yixuan Wei, Zheng Zhang, Stephen Lin, and Baining Guo.
\newblock Swin transformer: Hierarchical vision transformer using shifted windows.
\newblock In {\em Proceedings of the IEEE/CVF international conference on computer vision}, pages 10012--10022, 2021.

\bibitem{liu2022convnet}
Zhuang Liu, Hanzi Mao, Chao-Yuan Wu, Christoph Feichtenhofer, Trevor Darrell, and Saining Xie.
\newblock A convnet for the 2020s.
\newblock In {\em Proceedings of the IEEE/CVF Conference on Computer Vision and Pattern Recognition}, pages 11976--11986, 2022.

\bibitem{long2015fully}
Jonathan Long, Evan Shelhamer, and Trevor Darrell.
\newblock Fully convolutional networks for semantic segmentation.
\newblock In {\em Proceedings of the IEEE conference on computer vision and pattern recognition}, pages 3431--3440, 2015.

\bibitem{luo2021diffusion}
Shitong Luo and Wei Hu.
\newblock Diffusion probabilistic models for 3d point cloud generation.
\newblock In {\em Proceedings of the IEEE/CVF Conference on Computer Vision and Pattern Recognition}, pages 2837--2845, 2021.

\bibitem{nichol2021improved}
Alexander~Quinn Nichol and Prafulla Dhariwal.
\newblock Improved denoising diffusion probabilistic models.
\newblock In {\em International Conference on Machine Learning}, pages 8162--8171. PMLR, 2021.

\bibitem{paszke2017automatic}
Adam Paszke, Sam Gross, Soumith Chintala, Gregory Chanan, Edward Yang, Zachary DeVito, Zeming Lin, Alban Desmaison, Luca Antiga, and Adam Lerer.
\newblock Automatic differentiation in pytorch.
\newblock 2017.

\bibitem{philion2020lift}
Jonah Philion and Sanja Fidler.
\newblock Lift, splat, shoot: Encoding images from arbitrary camera rigs by implicitly unprojecting to 3d.
\newblock In {\em Computer Vision--ECCV 2020: 16th European Conference, Glasgow, UK, August 23--28, 2020, Proceedings, Part XIV 16}, pages 194--210. Springer, 2020.

\bibitem{poole2022dreamfusion}
Ben Poole, Ajay Jain, Jonathan~T Barron, and Ben Mildenhall.
\newblock Dreamfusion: Text-to-3d using 2d diffusion.
\newblock {\em arXiv preprint arXiv:2209.14988}, 2022.

\bibitem{radford2021learning}
Alec Radford, Jong~Wook Kim, Chris Hallacy, Aditya Ramesh, Gabriel Goh, Sandhini Agarwal, Girish Sastry, Amanda Askell, Pamela Mishkin, Jack Clark, et~al.
\newblock Learning transferable visual models from natural language supervision.
\newblock In {\em International conference on machine learning}, pages 8748--8763. PMLR, 2021.

\bibitem{ramesh2022hierarchical}
Aditya Ramesh, Prafulla Dhariwal, Alex Nichol, Casey Chu, and Mark Chen.
\newblock Hierarchical text-conditional image generation with clip latents.
\newblock {\em arXiv preprint arXiv:2204.06125}, 2022.

\bibitem{rezende2015variational}
Danilo Rezende and Shakir Mohamed.
\newblock Variational inference with normalizing flows.
\newblock In {\em International conference on machine learning}, pages 1530--1538. PMLR, 2015.

\bibitem{rombach2022high}
Robin Rombach, Andreas Blattmann, Dominik Lorenz, Patrick Esser, and Bj{\"o}rn Ommer.
\newblock High-resolution image synthesis with latent diffusion models.
\newblock In {\em Proceedings of the IEEE/CVF Conference on Computer Vision and Pattern Recognition}, pages 10684--10695, 2022.

\bibitem{ronneberger2015u}
Olaf Ronneberger, Philipp Fischer, and Thomas Brox.
\newblock U-net: Convolutional networks for biomedical image segmentation.
\newblock In {\em Medical Image Computing and Computer-Assisted Intervention--MICCAI 2015: 18th International Conference, Munich, Germany, October 5-9, 2015, Proceedings, Part III 18}, pages 234--241. Springer, 2015.

\bibitem{saharia2022palette}
Chitwan Saharia, William Chan, Huiwen Chang, Chris Lee, Jonathan Ho, Tim Salimans, David Fleet, and Mohammad Norouzi.
\newblock Palette: Image-to-image diffusion models.
\newblock In {\em ACM SIGGRAPH 2022 Conference Proceedings}, pages 1--10, 2022.

\bibitem{saharia2022photorealistic}
Chitwan Saharia, William Chan, Saurabh Saxena, Lala Li, Jay Whang, Emily Denton, Seyed Kamyar~Seyed Ghasemipour, Burcu~Karagol Ayan, S~Sara Mahdavi, Rapha~Gontijo Lopes, et~al.
\newblock Photorealistic text-to-image diffusion models with deep language understanding.
\newblock {\em arXiv preprint arXiv:2205.11487}, 2022.

\bibitem{sandler2018mobilenetv2}
Mark Sandler, Andrew Howard, Menglong Zhu, Andrey Zhmoginov, and Liang-Chieh Chen.
\newblock Mobilenetv2: Inverted residuals and linear bottlenecks.
\newblock In {\em Proceedings of the IEEE conference on computer vision and pattern recognition}, pages 4510--4520, 2018.

\bibitem{schuhmann2022laion}
Christoph Schuhmann, Romain Beaumont, Richard Vencu, Cade Gordon, Ross Wightman, Mehdi Cherti, Theo Coombes, Aarush Katta, Clayton Mullis, Mitchell Wortsman, et~al.
\newblock Laion-5b: An open large-scale dataset for training next generation image-text models.
\newblock {\em arXiv preprint arXiv:2210.08402}, 2022.

\bibitem{sohl2015deep}
Jascha Sohl-Dickstein, Eric Weiss, Niru Maheswaranathan, and Surya Ganguli.
\newblock Deep unsupervised learning using nonequilibrium thermodynamics.
\newblock In {\em International Conference on Machine Learning}, pages 2256--2265. PMLR, 2015.

\bibitem{song2019generative}
Yang Song and Stefano Ermon.
\newblock Generative modeling by estimating gradients of the data distribution.
\newblock {\em Advances in neural information processing systems}, 32, 2019.

\bibitem{van2017neural}
Aaron Van Den~Oord, Oriol Vinyals, et~al.
\newblock Neural discrete representation learning.
\newblock {\em Advances in neural information processing systems}, 30, 2017.

\bibitem{villegas2022phenaki}
Ruben Villegas, Mohammad Babaeizadeh, Pieter-Jan Kindermans, Hernan Moraldo, Han Zhang, Mohammad~Taghi Saffar, Santiago Castro, Julius Kunze, and Dumitru Erhan.
\newblock Phenaki: Variable length video generation from open domain textual description.
\newblock {\em arXiv preprint arXiv:2210.02399}, 2022.

\bibitem{wang2021towards}
Xintao Wang, Yu Li, Honglun Zhang, and Ying Shan.
\newblock Towards real-world blind face restoration with generative facial prior.
\newblock In {\em Proceedings of the IEEE/CVF Conference on Computer Vision and Pattern Recognition}, pages 9168--9178, 2021.

\bibitem{wang2022zero}
Yinhuai Wang, Jiwen Yu, and Jian Zhang.
\newblock Zero-shot image restoration using denoising diffusion null-space model.
\newblock {\em arXiv preprint arXiv:2212.00490}, 2022.

\bibitem{wightman2021resnet}
Ross Wightman, Hugo Touvron, and Herv{\'e} J{\'e}gou.
\newblock Resnet strikes back: An improved training procedure in timm.
\newblock {\em arXiv preprint arXiv:2110.00476}, 2021.

\bibitem{wolleb2022diffusion}
Julia Wolleb, Robin Sandk{\"u}hler, Florentin Bieder, Philippe Valmaggia, and Philippe~C Cattin.
\newblock Diffusion models for implicit image segmentation ensembles.
\newblock In {\em International Conference on Medical Imaging with Deep Learning}, pages 1336--1348. PMLR, 2022.

\bibitem{wu2022medsegdiff}
Junde Wu, Huihui Fang, Yu Zhang, Yehui Yang, and Yanwu Xu.
\newblock Medsegdiff: Medical image segmentation with diffusion probabilistic model.
\newblock {\em arXiv preprint arXiv:2211.00611}, 2022.

\bibitem{wu2023medsegdiff}
Junde Wu, Rao Fu, Huihui Fang, Yu Zhang, and Yanwu Xu.
\newblock Medsegdiff-v2: Diffusion based medical image segmentation with transformer.
\newblock {\em arXiv preprint arXiv:2301.11798}, 2023.

\bibitem{xiao2018unified}
Tete Xiao, Yingcheng Liu, Bolei Zhou, Yuning Jiang, and Jian Sun.
\newblock Unified perceptual parsing for scene understanding.
\newblock In {\em Proceedings of the European Conference on Computer Vision (ECCV)}, pages 418--434, 2018.

\bibitem{xie2021segformer}
Enze Xie, Wenhai Wang, Zhiding Yu, Anima Anandkumar, Jose~M Alvarez, and Ping Luo.
\newblock Segformer: Simple and efficient design for semantic segmentation with transformers.
\newblock {\em Advances in Neural Information Processing Systems}, 34:12077--12090, 2021.

\bibitem{yu2020context}
Changqian Yu, Jingbo Wang, Changxin Gao, Gang Yu, Chunhua Shen, and Nong Sang.
\newblock Context prior for scene segmentation.
\newblock In {\em Proceedings of the IEEE/CVF conference on computer vision and pattern recognition}, pages 12416--12425, 2020.

\bibitem{yuan2020object}
Yuhui Yuan, Xilin Chen, and Jingdong Wang.
\newblock Object-contextual representations for semantic segmentation.
\newblock In {\em Computer Vision--ECCV 2020: 16th European Conference, Glasgow, UK, August 23--28, 2020, Proceedings, Part VI 16}, pages 173--190. Springer, 2020.

\bibitem{zhang2021k}
Wenwei Zhang, Jiangmiao Pang, Kai Chen, and Chen~Change Loy.
\newblock K-net: Towards unified image segmentation.
\newblock {\em Advances in Neural Information Processing Systems}, 34:10326--10338, 2021.

\bibitem{zhao2018icnet}
Hengshuang Zhao, Xiaojuan Qi, Xiaoyong Shen, Jianping Shi, and Jiaya Jia.
\newblock Icnet for real-time semantic segmentation on high-resolution images.
\newblock In {\em Proceedings of the European conference on computer vision (ECCV)}, pages 405--420, 2018.

\bibitem{zhao2017pyramid}
Hengshuang Zhao, Jianping Shi, Xiaojuan Qi, Xiaogang Wang, and Jiaya Jia.
\newblock Pyramid scene parsing network.
\newblock In {\em Proceedings of the IEEE conference on computer vision and pattern recognition}, pages 2881--2890, 2017.

\bibitem{zheng2021rethinking}
Sixiao Zheng, Jiachen Lu, Hengshuang Zhao, Xiatian Zhu, Zekun Luo, Yabiao Wang, Yanwei Fu, Jianfeng Feng, Tao Xiang, Philip~HS Torr, et~al.
\newblock Rethinking semantic segmentation from a sequence-to-sequence perspective with transformers.
\newblock In {\em Proceedings of the IEEE/CVF conference on computer vision and pattern recognition}, pages 6881--6890, 2021.

\bibitem{zhou2017scene}
Bolei Zhou, Hang Zhao, Xavier Puig, Sanja Fidler, Adela Barriuso, and Antonio Torralba.
\newblock Scene parsing through ade20k dataset.
\newblock In {\em CVPR}, pages 633--641, 2017.

\end{thebibliography}
